%% file: main.tex
\pdfoutput=1

\documentclass[11pt]{article}

\usepackage{acl}

\usepackage{times}
\usepackage{latexsym}

\usepackage[T1]{fontenc}

\usepackage[utf8]{inputenc}

\usepackage{microtype}

\usepackage{inconsolata}

\input{math_commands.tex}

\usepackage{hyperref}
\usepackage{url}

\usepackage[normalem]{ulem}
\usepackage{caption}
\usepackage{subcaption}
\usepackage{graphicx}
\usepackage{wrapfig}
\usepackage{amsthm}
\usepackage{amsmath}
\usepackage{multirow}

\newtheorem{theorem}{Theorem}

\newtheorem{lemma}[theorem]{Lemma}
\newtheorem{proposition}{Proposition}
\newtheorem{observation}{Observation}
\usepackage{longtable}
\usepackage{booktabs}
\usepackage{xspace}
\usepackage{tabularray}
\usepackage{tabularx}
\usepackage[export]{adjustbox}

\title{\model{}:
Zero-Shot Extreme Length Generalization \\ for Large Language Models
}

\author{
Chi Han$^1$ \thanks{Work performed as an intern in Meta GenAI.},
Qifan Wang$^2$, Hao Peng$^1$, Wenhan Xiong$^3$, Yu Chen$^4$ \thanks{This work was done while the author was at Meta.}, Heng Ji$^1$, Sinong Wang$^3$ \\
\\
$^1$ University of Illinois Urbana-Champaign, $^2$ Meta,  $^3$ GenAI Meta,  $^4$ Anytime AI\\
$^1$\texttt{\{chihan3,haopeng,hengji\}@illinois.edu}, \\
$^{23}$\texttt{\{wqfcr,xwhan,sinongwang\}@meta.com}, \\
$^{4}$\texttt{ychen@anytime-ai.com}
}

\newcommand{\model}{LM-Infinite\xspace}
\usepackage{graphicx}

\begin{document}

\newcommand{\ensuretext}[1]{#1}
\newcommand{\marker}[2]{\ensuremath{^{\textsc{#1}}_{\textsc{#2}}}}
\newcommand{\draftcomment}[3]{\ensuretext{\textcolor{#3}{[#1 #2]}}}

\newcommand{\hao}[1]{\draftcomment{\marker{H}{P}}{#1}{purple}}
\NewDocumentCommand{\heng}{ mO{} }{\textcolor{red}{\textsuperscript{\textit{Heng}}\textsf{\textbf{\small[#1]}}}}
\NewDocumentCommand{\chihan}{ mO{} }{\textcolor{blue}{\textsuperscript{\textit{Chi}}{#1}}}
\newcommand{\hugo}[1]{{\color{red}{\textsuperscript{\textit{Hugo}} #1}}}


\maketitle

\begin{abstract}
    \input{body/0_abstract}
\end{abstract}

\input{body/1_introduction}

\input{figures/diagnosis}

\input{body/2_related}

\input{body/3_diagnosis}
\input{body/3_1_unseen_distances}
\input{figures/overview}
\input{body/3_2_too_many}
\input{body/3_3_implicit_positions}

\input{body/4_approach}

\input{figures/perplexity}
\input{tables/perplexity}

\input{body/4_1_principles}
\input{body/4_3_conceptual_model}

\input{figures/extreme}

\input{body/5_experiments}
\input{body/5_1_perplexity}

\input{figures/ablation}

\input{tables/downstream}
\input{tables/generation}

\input{body/5_2_ablation}
\input{body/5_3_downstream}
\input{figures/truncation_comparison}
\input{body/5_4_generation}

\input{body/6_conclusion}

\input{body/7_limitations}
\input{body/8_acknowledgement}

\bibliography{custom}

\appendix
\clearpage
\input{body/A_implementation}
\input{body/A_related_work}
\input{body/A_proof1}

\input{body/A_proof2}

\input{body/A_layer_positions}

\input{body/A_pseudo_dimension}
\input{body/A_efficiency}
\input{body/A_example_generation}

\end{document}

%% file: math_commands.tex
\usepackage{amsmath,amsfonts,bm}

\def\eqref#1{equation~\ref{#1}}

\def\1{\bm{1}}

\DeclareMathAlphabet{\mathsfit}{\encodingdefault}{\sfdefault}{m}{sl}
\SetMathAlphabet{\mathsfit}{bold}{\encodingdefault}{\sfdefault}{bx}{n}



%% file: body/0_abstract.tex
Today's large language models (LLMs) typically train on short text segments (e.g., <4K tokens) due to the quadratic complexity of their Transformer architectures. As a result, their performance suffers drastically on inputs longer than those encountered during training, substantially limiting their applications in real-world tasks involving long contexts such as encoding scientific articles, code repositories, or long dialogues. Through both theoretical analysis and empirical investigation, this work identifies three major factors contributing to this length generalization failure. Our theoretical analysis reveals that commonly used techniques like using a sliding-window attention pattern or relative positional encodings are inadequate to address them. Answering these challenges, we propose \model, a simple and effective method for enhancing LLMs' capabilities of handling long contexts. \model is highly flexible and can be used with most modern LLMs off-the-shelf.
\emph{Without any parameter updates}, it allows LLMs pre-trained with 2K or 4K-long segments to generalize to up to 200M length inputs while retaining perplexity.
It also improves performance on downstream tasks such as Passkey Retrieval and Qasper in the zero-shot setting. \model brings substantial efficiency improvements: it achieves 2.7$\times$ decoding speed up and 7.5$\times$ memory saving over the original model.
Our codes are released at \url{https://github.com/Glaciohound/LM-Infinite}.

%% file: body/1_introduction.tex
\section{Introduction}
\label{sec:introduction}

Large language models (LLMs) have recently advanced the state-of-the-art across various natural language processing tasks. They typically train on text segments of fewer than 4K tokens~\citep{touvron2023llama2, MosaicML2023Introducing}, primarily due to the computational overhead quadratic in the input lengths of their Transformer architectures. As a result, they face challenges in generalization to inputs that are excessively longer than what they are trained on and suffer substantial deterioration 
 in their performance~\cite {tworkowski2023focused, chen2023extending}. This limits their applicability in tasks that require long-range contexts, such as encoding scientific articles, source code repository generation, or long-context dialogues.

Extensive efforts have been devoted to addressing this length generalization challenge.
Relative positional encodings such as RoPE~\citep{su2021roformer} and Alibi~\citep{press2021train} have been widely adopted by state-of-the-art LLMs, which calculate attention based on inter-token distance instead of absolute positions, hoping to avoid model failures due to unseen absolute position embeddings.
Moreover, although applying a sliding-window attention pattern on the Transformer architecture can reduce the memory overhead~\citep{beltagy2020longformer, ding2023longnet, zaheer2020big}, they are not directly applicable to pre-trained models for length generalization without further training.
Through both theoretical analysis and empirical investigation, \S\ref{sec:diagnosis} pinpoints three primary factors underlying the length generalization failures:
(1) the challenge of handling unseen distances among tokens, 
(2) the difficulty of attending to unseen numbers of tokens, 
and (3) implicitly encoded absolute positional information in initial tokens.
These challenges can make LLMs' computational features, such as attention logits and hidden vectors, deviate from the training distribution,
leading to failures of length generalization.
Existing techniques fall short of addressing these underlying issues.

Answering these challenges, we propose \model, a simple and effective method to enhance Transformer LLMs' capabilities for modeling long contexts\emph{without parameter updates}. 
\model consists of two major components designed to alleviate the three factors above. (1) a $\boldsymbol{\Lambda}$-shaped attention mask and (2) a ceiling on attention distances. 
The former forces the model to attend to only the beginning of the sequence and the most recent tokens within a pre-defined window, ignoring the rest.
The latter component caps the relative distance values to the maximum the model has seen during training. It can also optionally re-introduce top-$k$ tokens in the middle to achieve better performance in some downstream tasks.
\model{} is highly flexible and applies to any off-the-shelf LLMs that use relative positional encoding and does \emph{not} require any finetuning.

Our experiments thoroughly evaluate \model on a variety of tasks and LLMs.
On ArXiv (academic papers) and OpenWebText2 (Reddit posts)
\model facilitates  \emph{zero-shot} generalization for a wide range of LLMs to texts up to 200M tokens, retaining the language modeling perplexity and generation quality.
\emph{Without any parameter updates}, \model{} improves scores compared with the original model and truncation baselines on downstream tasks including Passkey Retrieval~\citep{mohtashami2023landmark} and Qasper~\citep{dasigi2021dataset}, which are two established benchmarks for long-context evaluation. We observe a 37.2\% gain on Passkey Retrieval and a 1.2\% gain on Qasper in the zero-shot setting.
\model also brings substantial efficiency improvements: it achieves 2.7$\times$ decoding speed up and 7.5$\times$ GPU memory saving over the original LLMs.

%% file: figures/diagnosis.tex

\begin{figure*}[t!]
\centering
\includegraphics[width=\textwidth]{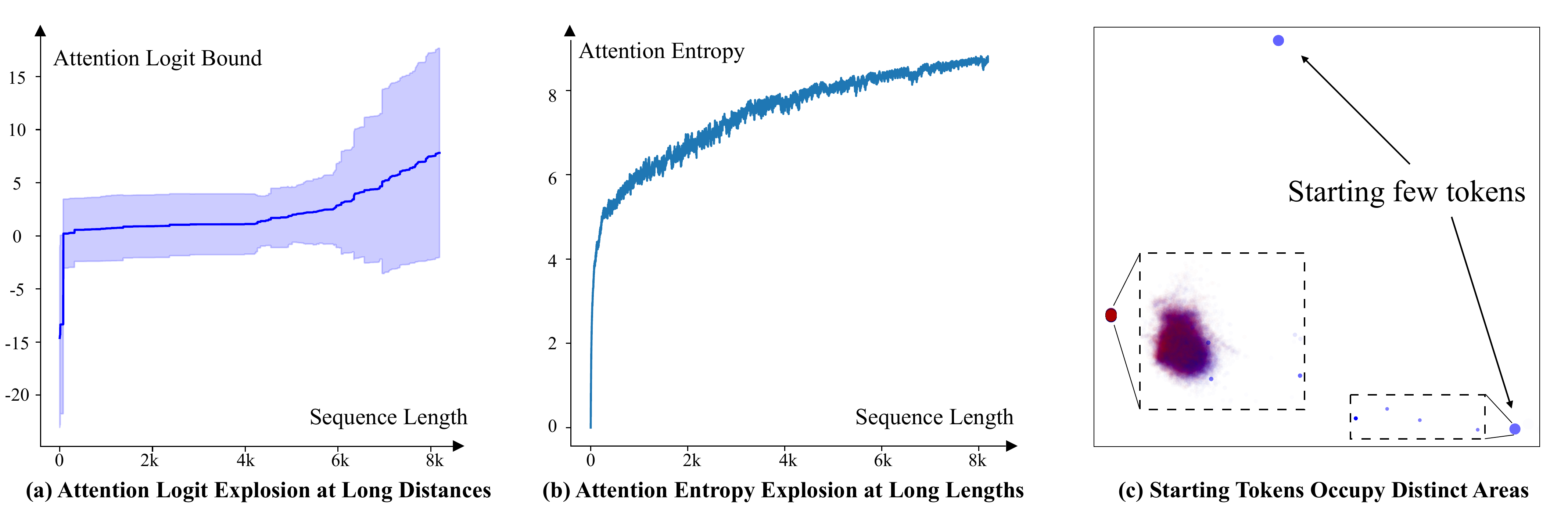}

\caption{We identify three factors underlying the length generalization failure in LLMs in \S\ref{sec:diagnosis}. (a) Factor 1: Unseen distances between tokens cause attention logits to explode. (b) Factor 2: An unseen number of tokens can cause attention entropy to increase beyond the training range as the length increases. (c) Factor 3: Starting few tokens occupy a distinct feature region and should not be discarded. The two blue regions at the upper center and lower right correspond to the initial tokens that are highly concentrated but also very far from later tokens. The lower-left region contains the thousands of overlapping dots corresponding to the later tokens.
}
\label{fig:diagnosis}
\end{figure*}

%% file: body/2_related.tex
\section{Background and Related Work}
\label{sec:related}

\subsection{Relative Positional Encodings}
The traditional absolute positional encodings provide the absolute position information, usually with the help of a sequence of vectors called \textit{position embeddings}~\citep{vaswani2017attention, kenton2019bert, ke2020rethinking}. They, however, have trouble when the model encounters unseen positions in inputs longer than the training length.
Relative positional encodings aim to address the limitations of previous-generation positional encoding methods and consider the relative distances between tokens instead of the absolute positions.
Examples include a learned attention logit bias in T5~\citep{raffel2020exploring}, Transformer-XL~\cite{dai2019Transformer}, Skyformer~\cite{skyformer2021}, Sketching~\cite{Sketching2022} and Sandwich~\citep{chi2023dissecting}, a fixed linear attention decay~\cite{press2021train}, and rotating query and key sequences based on distances such as RoPE~\citep{su2021roformer, longchat2023}, CAPE~\cite{likhomanenko2021cape} and XPos~\citep{sun2022length, ding2023longnet}. 
Despite some promising empirical evidence, length generalization failures are still widely observed when directly applied to large language models~\citep{superhot}. 
In what follows, we briefly discuss two widely used relative positional encoding methods.
They lay out the necessary context for our onward discussion and experiments.

\paragraph{Rotary Position Embedding (RoPE; \citealp{su2021roformer})}
It rotates the key and query vectors based on positions before computing the inner product. Specifically, each vector $\mathbf{x}$ (either key or query) is split into pairs of elements $\{(x_0, x_1), (x_2, x_3), \cdots\}$, with each pair interpreted as a 2-dimensional vector. 
RoPE then rotates the vector $(x_a, x_{a+1})$ of token $i$ with angle $\theta_{a, i}=i\omega_a$, where $\omega_a$ is the rotating speed associated with dimension pair $(a, a+1)$. After rotation, the 2-D vector becomes $\begin{pmatrix}
\cos i\omega_a & -\sin i \omega_a\\
\sin i\omega_a & \cos i\omega_a
\end{pmatrix} \begin{pmatrix}
    x_i \\ x_{i+1}
\end{pmatrix}$. 
They show that the inner product between rotated query $\mathbf{q}_i$ and rotated key $\mathbf{k}_j$ is solely determined by $\mathbf{q}_i$, $\mathbf{k}_j$, and their relative distance $i-j$. We always have $i\geq j$ due to the causal attention mask.

\paragraph{AliBi~\cite{press2021train}} It offsets all attention logits between tokens $i, j$ by a linear term $-m(i-j)$ and become  $\mathbf{q}_i^\top \mathbf{k}_j - m(i-j)$. To this end, the MPT-7B codes implement an offset matrix as an additive term in attention logits.

\subsection{Efforts Towards Length Generalization}
In light of generalization failures observed in LLMs, one straightforward solution is to finetune LLMs on longer text sequences~\citep{chen2023extending, tworkowski2023focused, tao2023frustratingly, kiyono2021shape, anil2022exploring}. These approaches do not address the underlying causes of length generalization failures and require massive training resources.
Other solutions propose to grant LLMs access to longer contexts without really reading them in full~\citep{zhou2023recurrentgpt, bueno2022induced, mohtashami2023landmark, yang-etal-2023-doc}. Augmenting LLMs with retrieval-based memories~\cite{wu2021memorizing, guu2020retrieval, borgeaud2022improving, khandelwal2019generalization, kaiser2016learning, yogatama2021adaptive} also make LLMs applicable to a large database. These designs, however, usually need finetuning and are not directly compatible with the existing LLMs. Our work, in contrast, facilitates zero-shot length generalization. Another similar work \cite{ratner2023parallel} increases context length with attention patterns without further training. However, it is limited to the in-context learning setting.

%% file: body/3_diagnosis.tex
\section{Why do Transformer LLMs Fail to Generalize to Long Contexts?}
\label{sec:diagnosis}

Through a series of theoretical and experimental investigations, this section aims to identify the potential causes underlying current LLMs' failure in length generalization.
Our discussion assumes Transformer-based LLMs that use relative positional encodings, as this design is widely adopted in today's LLMs~\cite{touvron2023llama2, MosaicML2023Introducing}. We use Llama-2~\citep{touvron2023llama2}, which is pre-trained with 4K-length segments, for investigation.
On sequences longer than the training length, we will show that the \textit{unseen inter-token distances}, the \textit{increasing number of attended tokens}, and the \textit{implicitly encoded position information of the starting tokens} can all make certain computational features out of the training distribution. As deep models can be sensitive to input distribution shifts, these factors need to be handled for LLMs to generalize to unseen lengths.

%% file: body/3_1_unseen_distances.tex
\paragraph{Factor 1: challenges in handling unseen distances among tokens}

With relative positional encoding, the impact of positions on the attention weight between two tokens depends solely on their relative distance.
As the sequence length grows exceedingly long, some distance values will surpass those seen during training. 
We make the following informal theoretical claim:
\begin{theorem}(Informal)
\label{thm:attention_logit_explosion_informal}
    For an attention mechanism using relative positional encoding, the attention logits must explode to infinities to differentiate previously unseen distances apart as the sequence length increases.
\end{theorem}

\noindent The formal theorem and its proof can be found in Appendix~\ref{sec:proof_explosion}. We also empirically verify this on Llama-2 on the ArXiv dataset truncated down to 8K length. We extract the attention logits of all attention heads and their maximum attention logits on different sequence lengths in Figure~\ref{fig:diagnosis}(a). It shows the average and variance among attention heads. We see that the attention logits increase to substantially larger values when the sequence length exceeds the training length of 4K.
To mitigate this issue, we conjecture that \textbf{it may help to cap the relative distance values to the maximum that the model has seen during training (i.e.,  a distance ceiling)}.
However, as we will see from the proposition below, addressing logit explosion leads to another challenge.

%% file: figures/overview.tex

\begin{figure*}[t!]
\centering
\includegraphics[width=\textwidth]{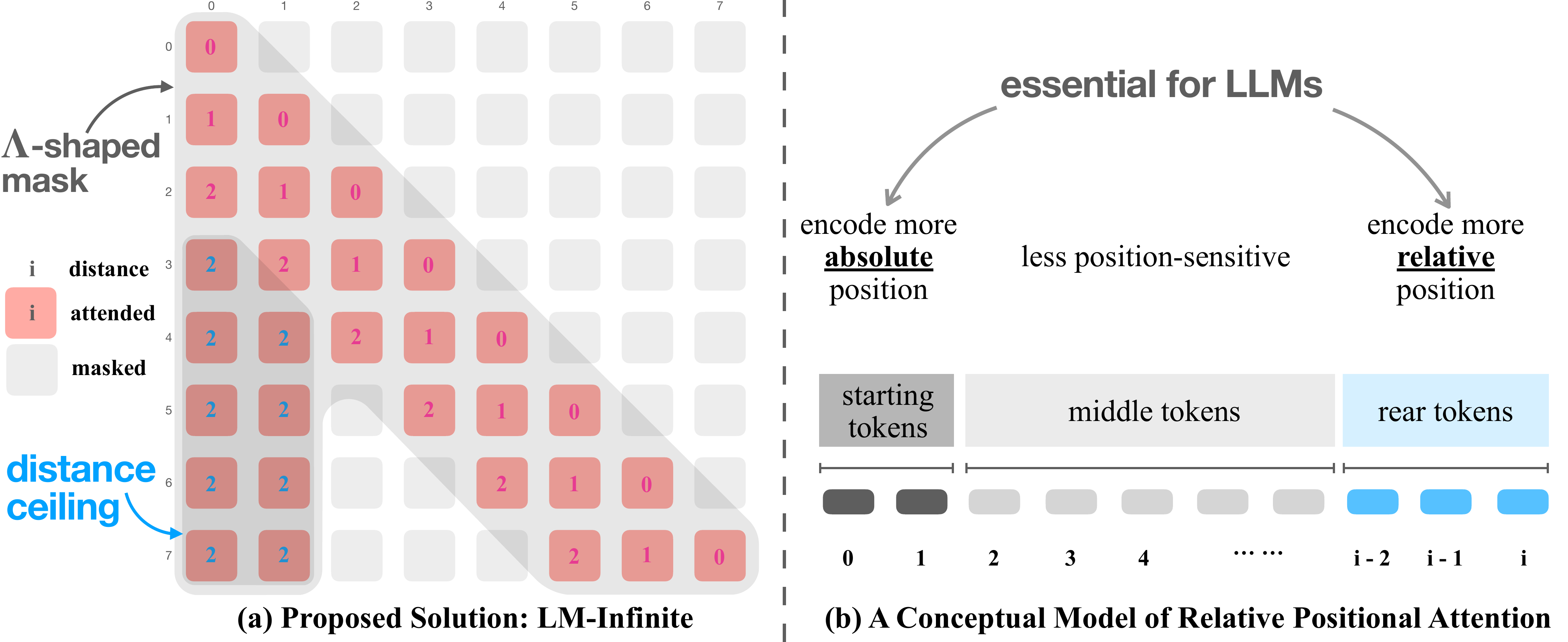}

\caption{(a) \model{} is a plug-and-play solution for various LLMs, consisting of a $\mathbf{\Lambda}$-shaped mask and a distance ceiling during attention. For clarity, this figure shows a toy scenario where $L_\text{pre-train}$ and $n_\text{starting}$ are both 2. (b) We also provide a conceptual model for understanding how relative position encoding works.}
\label{fig:overview}
\end{figure*}

%% file: body/3_2_too_many.tex
\paragraph{Factor 2: attending to unseen numbers of tokens}

On longer sequences, tokens at later positions must distribute attention weights across a larger context.
We then make the following claim that, if attention logits are bounded but the number of tokens to attend to is not limited, it can cause the attention entropy to increase beyond the training range:
\begin{proposition}
\label{thm:entropy_explosion_informal}
    If the attention logits are bounded, as the sequence becomes longer, the attention entropy grows to infinity.
\end{proposition}

\noindent A formal statement as well as the proof can be found in Appendix~\ref{sec:entropy_explosion}.
This conclusion is further empirically verified by plotting the attention entropy against context lengths in Figure~\ref{fig:diagnosis}(b). 
The curve shows an ever-increasing attention entropy.
This trend, although increasing logarithmically, is still harming the LLMs' performance, as we will illustrate in the ablation study in \S\ref{sec:ablation} and Figure~\ref{fig:ablation}.
This suggests that \textbf{we should bound the attention context size} to ensure that the attention entropy stays in seen ranges during pre-training and avoid degenerated outputs.
A simple windowed attention, where each token only attends to the nearest tokens within a distance, might handle factors 1 and 2.
This is similar to the block-diagonal attention mask used in XPos~\citep{sun2022length} and Longformer~\citep{beltagy2020longformer}. However, as we will show in the next paragraph, this introduces another factor that can also fail LLMs.

%% file: body/3_3_implicit_positions.tex
\paragraph{Factor 3: starting tokens occupy a distinct feature space}

Perhaps counter-intuitively:
\begin{observation}
Even without explicit absolute positional embeddings, attention outputs of the first few tokens can occupy a distinct representational space compared to other positions. Therefore, when passed to later layers, these starting tokens have distinct value vectors from their lower-layer outputs.
\end{observation}
\noindent This follows from Theorem 1 in \citet{kazemnejad2023impact}, which proves that the absolute positions can be implicitly encoded in the outputs of tokens of a single attention layer, even \emph{without} positional encodings. In their construction, the starting tokens' signals are the strongest and easiest to distinguish from other tokens. As relative positional encoding is strictly more expressive than no positional encoding setting in \citet{kazemnejad2023impact} (e.g., by letting all distances have the same attention function), the same conclusion applies to relative positional encoding as well.

As an empirical verification, we take the hidden states output by the second layer of Llama-2 and plot a Principal Component Analysis (PCA) projection into a 2-d plane in Figure~\ref{fig:diagnosis}(c). More figures for other layers can be found in \S\ref{sec:layer_positions}. The dots correspond to the first 4096 tokens in 32 sequences, with blue ones corresponding to the initial tokens and red tokens being the tail ones.
The two blue regions at the upper center and lower right correspond to the initial highly concentrated tokens (whose positions are around 0$\sim$25) and are very far from later tokens. The lower-left region contains the remaining overlapping tokens in a sequence (zoomed in to another box).
The plot shows that the vector representations of the initial tokens concentrate on regions in the feature space that are distinct from the remaining tokens.
This fresh finding reveals a fundamental flaw of the sliding-window attention pattern, which restricts the attention to the most recent tokens within a predefined window size, a widely adopted baseline recently~\citep{beltagy2020longformer, ding2023longnet, zaheer2020big}.
As attention is essentially a weighted average over the value vectors, 
sliding-window attention discards the initial tokens, keeping the attention output from reaching the regions that value vectors of the initial tokens occupy. 
This enforces the model to handle a different region during the computation, introducing additional generalization challenges. 
As a straightforward solution to this issue, \textbf{the initial tokens need to be kept in the attention computation.}

%% file: body/4_approach.tex
\section{Our proposal: \model{}}
\label{sec:approach}

%% file: figures/perplexity.tex

\begin{figure*}[t!]
\centering
\includegraphics[width=\textwidth]{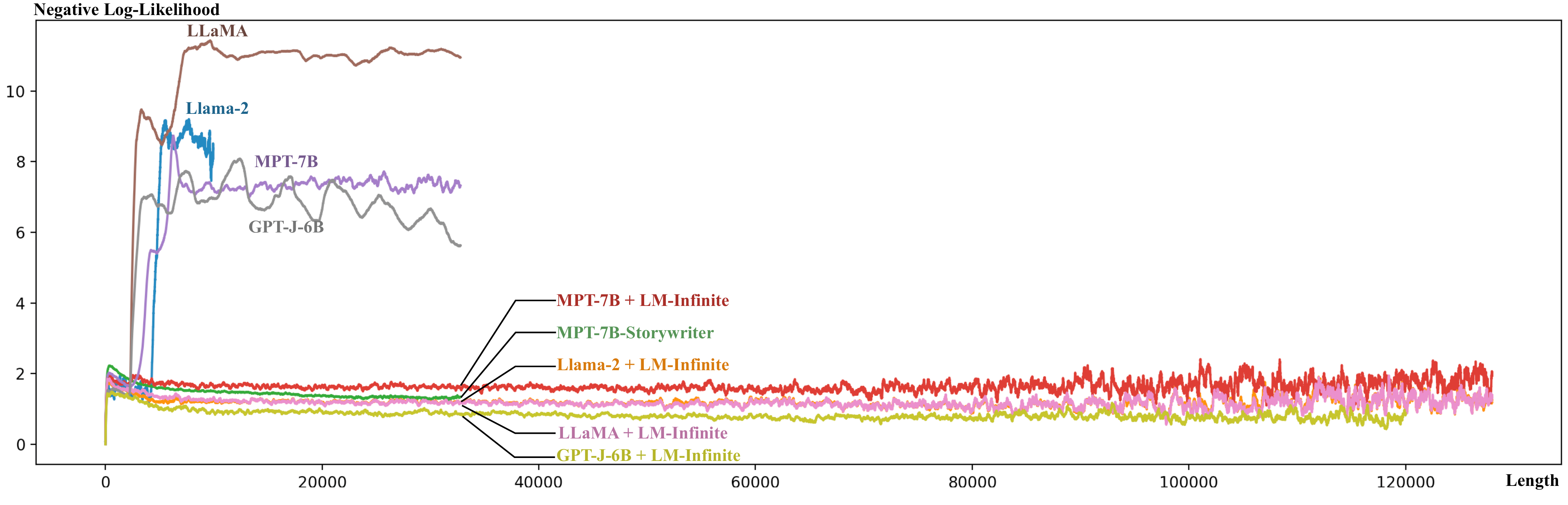}

\caption{\model{} flattens the negative log-likelihood (NLL) curves of various LLMs on ArXiv dataset without any parameter updates. The trends are similar to MPT-7B-Storywriter, an explicitly fine-tuned LLM. Llama-2 outputs NaN values on long sequences so the curve is relatively shorter. }
\label{fig:perplexity}
\end{figure*}

%% file: tables/perplexity.tex
\begin{table*}[t!]
\centering
\linespread{1}


\begin{adjustbox}{max width=\textwidth}
\begin{tabular}{@{} lc ccccc m{0.01em} cccc @{}}
\toprule

& & \multicolumn{5}{c}{\textbf{ArXiv}} && \multicolumn{4}{c}{\textbf{OpenWebText2}} \\
\cline{3-7}
\cline{9-12}

\textbf{Model} & $L_\text{pretrain}$ & \phantom{0}2K & \phantom{0}4K & \phantom{0}8K & 16K & 32K && \phantom{0}2K & \phantom{0}4K & \phantom{0}8K & 16K \\

\midrule
\multicolumn{12}{@{}l}{\textit{Long-context Training/Finetuning}}\\
\quad Sandwich & 512 & \phantom{0}5.0 & \phantom{0}5.2 & 5.3 & - & - && 23.3 & 23.8 & 24.7 & -  \\
\quad XPos & 1K & 21.6 & 20.7 & - & - & - && - & - & - & - \\
\quad LongLLaMA & 8K & \phantom{0}8.2 & \phantom{0}7.4 & - & 6.9 & - && - & - & - & - \\
\quad MPT-7B-SW & 65K & \phantom{0}6.5 & \phantom{0}5.4 & 4.3 & 4.4 & 3.6 && \phantom{0}9.8 & 10.9 & \phantom{0}6.6 & \textbf{5.1} \\

\midrule
\multicolumn{12}{@{}l}{\textit{Vanilla}} \\
\quad MPT-7B & 4K & 5.5 & 2.5$\times 10^2$ & 1.1$\times 10^3$ & 1.7$\times 10^3$ & 1.6$\times 10^3$ && 8.3 & 1.3$\times 10^2$ & 1.9$\times 10^2$ & 1.3$\times 10^2$ \\
\quad LLaMA & 2K & 3.8 & 1.0$\times 10^4$ & 6.0$\times 10^4$ & 6.8$\times 10^4$ & 4.9$\times 10^4$ && 6.2 & 6.6$\times 10^3$ & 4.6$\times 10^5$ & 4.4$\times 10^4$ \\
\quad GPT-J-6B &  2K & 3.9 & 1.3$\times 10^3$ & 1.0$\times 10^3$ & 1.6$\times 10^3$ & 2.8$\times 10^2$ &&  8.8 & 7.5$\times 10^2$ & 1.3$\times 10^3$ & 1.8$\times 10^3$  \\
\quad Llama-2 & 4K & \textbf{3.4} & 3.8 & 8.5$\times 10^3$ & NaN & NaN && 6.2 & 5.8 & 6.5$\times 10^3$ & NaN \\

\midrule
\multicolumn{12}{@{}l}{\textit{\model}}\\
\quad MPT-7B &  4K & 5.7 & 6.8 & 5.8 & 6.0 & 4.6 && 8.5 & 12.2 & 8.5 & 8.9 \\
\quad LLaMA &  2K & 4.4 & 4.5 & 3.7 & 4.2 & \textbf{1.0} && 6.3 & \phantom{0}6.1 & 9.5 & 7.0 \\
\quad GPT-J-6B &  2K & 3.8 & \textbf{3.1} & \textbf{3.0} & \textbf{3.1} & 2.1 && 8.8 & \phantom{0}8.5 & \textbf{6.5} & 7.4 \\
\quad Llama-2 &  4K & 4.3 & 3.6 & 3.3 & 4.2 & 6.5 && \textbf{6.1} & \phantom{0}\textbf{5.3} & 8.3 & 8.2 \\
\bottomrule

\end{tabular}
\end{adjustbox}

\caption{
Perplexity on ArXiv and OpenWebText2 test split. LLMs with \model{} achieve the highest perplexity on 7 out of 9 columns while requiring no parameter updates. $L_\text{pretrain}$ indicates the lengths of the text segments that the models are trained on.}
\label{tab:perplexity}
\end{table*}

%% file: body/4_1_principles.tex
Inspired by the analyses and take-away messages in the previous section, we propose \model to achieve zero-shot length generalization for LLMs. An overview of \model{} is illustrated in Figure~\ref{fig:overview}(a). This simple solution consists of two components: a $\mathbf{\Lambda}$-shaped attention mask and a distance ceiling. Besides, re-introducing the middle top-$k$ tokens is optional for enhanced downstream performance.

\paragraph{$\mathbf{\Lambda}$-shaped attention mask} It contains two attention spans: the \textit{starting} one allows each token to attend to the first $n_\text{starting}$ tokens if they come before the current one; the ending one allows each token to attend to most recent $L_\text{pretrain}$ tokens. $L_\text{pretrain}$ is the maximum length during training.
Other tokens are ignored.
In ablation studies in \S\ref{sec:evaluation_details}, we find that choosing $n_\text{starting}\in[5, 100]$ generally achieves equally good performance.
Note that $n_\text{starting}=0$ (i.e., attending only to the most recent tokens) substantially hurts the performance.
This resolves Factors 2 and 3 in \S\ref{sec:diagnosis} by both limiting the number of tokens under attention and ensuring the starting few tokens are attended.

\paragraph{Distance ceiling} \model further bounds  the ``effective distance'' to $L_\text{pretrain}$. This only affects the starting few tokens when attended by tokens at later positions. 
Specifically, in relative positional encoding, the original attention logit is $w(\mathbf{q}, \mathbf{k}, d)$, where $d$ is the distance between two tokens. Now we modify it as $w(\mathbf{q}, \mathbf{k}, d'))$ where $d'=\min(d, L_\text{pretrain})$. Figure~\ref{fig:overview}(a) shows an illustrative example where the distance ceiling is $L_\text{pretrain}=2$.
This addresses Factor 1 in \S\ref{sec:diagnosis} by bounding the distance value in attention calculation.
\paragraph{Optionally attending to top-$k$ tokens in the middle}
\model can optionally attend to $k$ tokens in the middle with the largest attention logits.
This is particularly useful in downstream tasks where information in the middle tokens matters (\S\ref{sec:downstream}). 
Here the $k$ tokens are selected independently for each attention head in layers higher than $h$-th layer, and have an attention distance of $d=\frac{1}{2}L_\text{pre-train}$. These hyperparameters are selected based on a held-out Passkey Retrieval validation set, where we set $k=5$ and $h=5$, with more details in Appendix~\ref{sec:evaluation_details}. Our selection of $k$ and $h$ does not depend on task-specific tuning, and in our experiments, we apply this same set of hyperparameters in all other downstream tasks and achieve consistent improvements.
These intermediate tokens do not hurt performance. Rather, in the evaluation of downstream tasks in \S\ref{sec:downstream}, intermediate tokens are more useful and selectively attending to top-$k$ tokens brings substantial performance improvements with little impact on the efficiency.
For LLM generation and inference, however, we find the intermediate tokens \emph{unnecessary} to attend to for \model to achieve good perplexity or generation quality. 

\model's $\mathbf{\Lambda}$-shaped mask is conceptually similar to the attention patterns derived from heuristics~\citep{beltagy2020longformer, ding2023longnet, zaheer2020big}.
However, we formally show in \S\ref{sec:diagnosis} Factor 3 that these previous approaches theoretically cannot generalize to unseen lengths but require parameter updates.
This inherent limitation motivates the other two components in \model to achieve zero-shot length generalization.

\paragraph{Implementation details}
\model{} is applicable in all Transformer models with relative positional encoding. One only needs to replace the attention function in each Transformer layer with \model without any parameter updates.
The $\Lambda$-shaped attention mask is relatively straightforward to implement. In RoPE, attention logits in the ending attention span follow the original calculation. In the starting attention span (excluding its overlap with the ending span), we keep all $\mathbf{k}$ vectors unrotated and rotate all $\mathbf{q}$ vectors to a fixed distance $L_\text{pretrain}$. Then the logits in two spans can be composed.
Augmenting AliBi with \model{} is also straightforward. We simply clip the offset matrix with a minimum value of $-|mL_\text{pretrain}|$ and apply the $\mathbf{\Lambda}$-shaped attention mask.

%% file: body/4_3_conceptual_model.tex
\paragraph{Discussion.}
In Figure~\ref{fig:overview}(b), we show a conceptual model of how relative positional encoding functions. This conceptual model reflects the design choices of \model. In this conceptual model, a long context can be roughly partitioned into 3 parts:
The \textit{starting tokens} encode strong absolute position information (Factor 3). As explained in \S\ref{sec:diagnosis}, they are essential to attention to because their features occupy a distinct region in the feature space. As attention is essentially a weighted average over $\mathbf{v}_i$ vectors, without the starting few tokens, the attention output can not reach regions that $\mathbf{v}_i$ vectors of the initial tokens occupy. 
The \textit{rear tokens} provides primarily their relative positions to the final tokens. Their importance probably arises from the ``recency bias''~\citep{peysakhovich2023attention} learned by LLMs during pre-training.
The \textit{middle tokens} encode less position-sensitive information. As analyzed in Factor 2, including too many intermediate tokens does more harm than good to length generalization.

%% file: figures/extreme.tex
\begin{figure}[t!]
\centering
\includegraphics[width=0.5\textwidth]{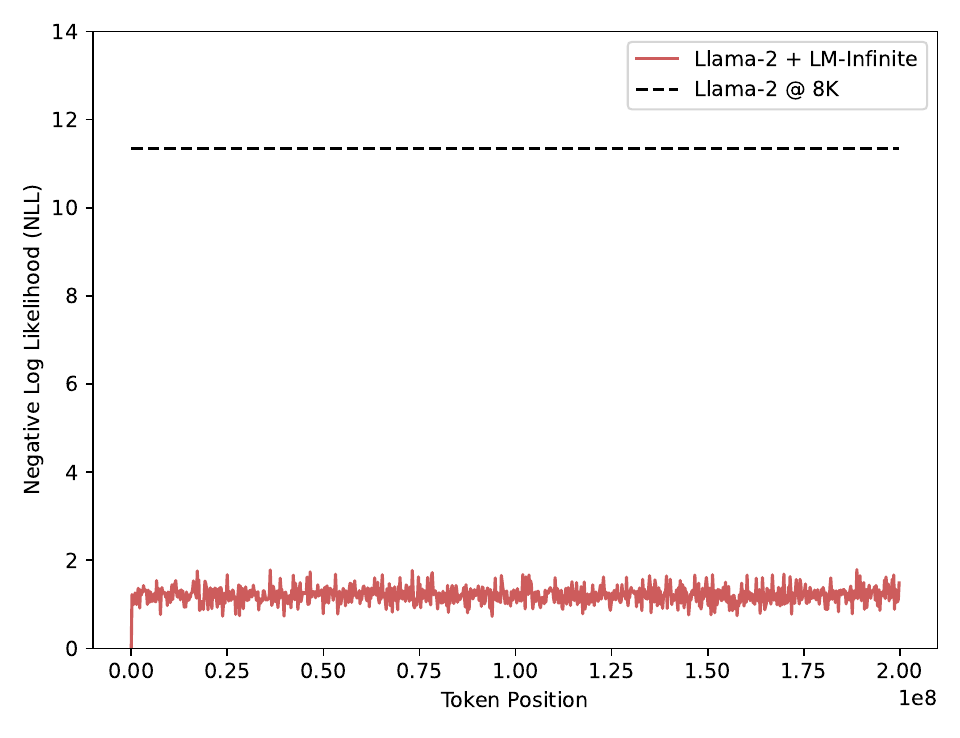}

\caption{\model generalizes Llama-2 to an extreme length of 200M. The dashed line is the NLL at 8K length of the vanilla Llama-2 model.}
\label{fig:extreme}
\end{figure}

%% file: body/5_experiments.tex
\section{Evaluation}
\label{sec:experiments}

We evaluate \model with LLaMA-7B~\citep{touvron2023llama}, Llama-2-7b~\citep{touvron2023llama2}, MPT-7B~\citep{MosaicML2023Introducing}, and GPT-J-6B~\citep{gpt-j}. LLaMA-7B and GPT-J-6B are pre-trained with 2K lengths and the other models are pre-trained with 4K lengths. LLaMA, Llama-2, and GPT-J use RoPE encoding, and MPT-7B uses Alibi encoding.
MPT-7B-Storywriter (fine-tuned on long sequences) is used as one of the baselines.

%% file: body/5_1_perplexity.tex
\subsection{Language Modeling with Extremely Long Context}
\label{sec:perplexity}
We use ArXiv and OpenWebText2 corpora from the Pile dataset~\citep{pile}, which contain preprint papers from ArXiv and Reddit submissions, respectively.
We evaluate with negative log-likelihood (NLL) and perplexity (exponential of NLL).
Figure~\ref{fig:perplexity} plots the NLL curves on the ArXiv dataset.
Here, we break down the models' perplexity performance by positions so that the curve shows the NLL that the model achieves around that specific position, averaged across all evaluated sequences.
Llama-2 outputs NaN probabilities on sequences that are slightly longer than 10K, thereby its shorter curve. 
All vanilla models run out of memory at $\sim$32K lengths.\footnote{We run on a single A100 GPU with 80GB GPU memory.} The baselines' NLL quickly blows up when the tested sequences are longer than what they train on. 
With \model{}, all models can generalize to sequences that are substantially longer than the lengths they are trained on, retaining the NLL performance. 
This validates our length failure factor analyses in \S\ref{fig:diagnosis}.
The longer ends of curves have larger variances because of fewer documents of those lengths.
In Figure~\ref{fig:extreme}, we further evaluate \model + Llama-2 on a sequence of \textbf{200M} tokens, which is constructed by sampling with replacement from the ArXiv dataset and concatenating all data. \model shows the ability to remain stably low log-perplexity level over extreme lengths.

Table~\ref{tab:perplexity} summarizes the perplexity performance at a few milestone lengths (2K, 4K, 8K, 16K, and 32K) on ArXiv and OpenWebText2, which shows a similar trend. OpenWebText2 has very few data instances over a length of 32K, so we omit the column. 
With \model{}, all models can generalize to unseen lengths, and \model achieves best perplexity in 7 out of 9 cases. On LLaMA + LM-Infinite, the perplexity decreases as length increases and position becomes larger.
Surprisingly, \emph{without any parameter update}, \model outperforms many strong baselines that are trained on substantially longer text segments.
As a direct comparison, MPT-7B+\model{} achieves only slightly worse performance than its fine-tuned counterpart, MPT-7B-Storywriter. This confirms that \model{} is a promising alternative to resource-consuming fine-tuning.

%% file: figures/ablation.tex
\begin{figure}[t!]
\centering
\includegraphics[width=0.5\textwidth]{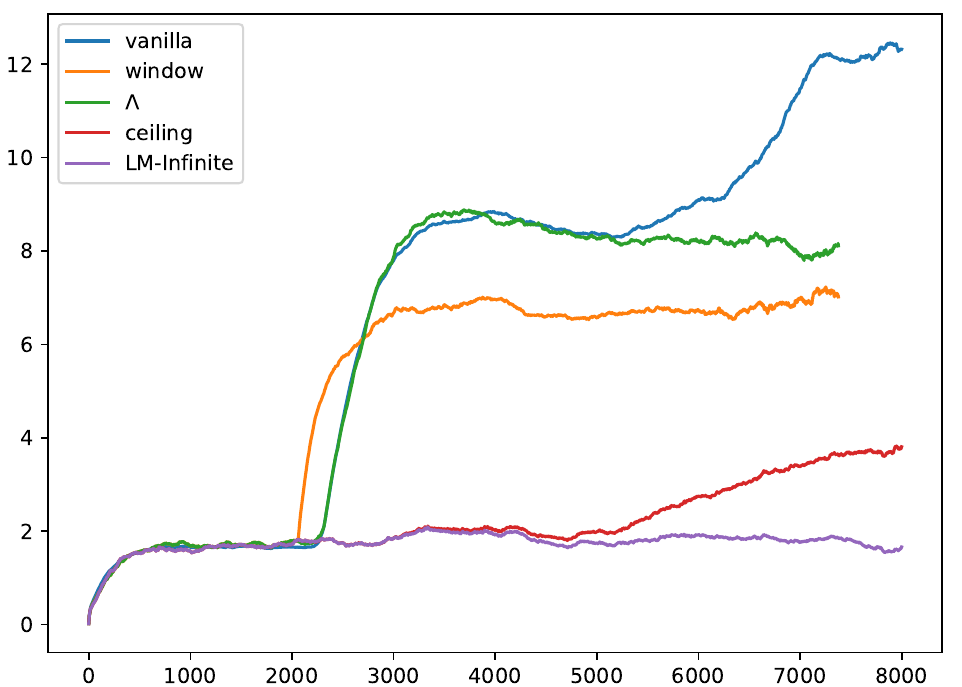}

\caption{Ablation study on LLaMA in \S\ref{sec:ablation}. x-axis is token position and y-axis is negative log-likelihood (NLL).
The vanilla model (vanilla), using a windowed attention (window), using only a $\Lambda$-shaped attention mask ($\Lambda$), and only the ceiling the distance value (ceiling) all more or less suffer from perplexity explosion. Only \model{} can retrain the performance while generalizing to unseen lengths.}
\label{fig:ablation}
\end{figure}

%% file: tables/downstream.tex
\begin{table*}[t]
\centering
\linespread{1}

\begin{tabular}{l cccccc c}
\toprule
& \multicolumn{6}{c}{\textbf{Passkey Retrieval}} & \textbf{Qasper} \\
\cline{2-7}
\textbf{Model} & 6K & 8K & 10K & 12K & 16K & average & \\
\midrule
\textbf{Original} & \phantom{0}0.0 & \phantom{0}0.0 & \phantom{0}0.0 & \phantom{0}0.0 & \phantom{0}0.0 & \phantom{0}0.0 & \phantom{0}1.2 \\
\textbf{Truncated} & 66.0 & 55.3 & 38.8 & 32.8 & 27.3 & 44.0 & 30.1 \\
\textbf{\model} & \textbf{70.3} & \textbf{90.8} & \textbf{86.5} & \textbf{79.3} & \textbf{79.1} & \textbf{81.2} & \textbf{31.3} \\
\bottomrule
\end{tabular}
\caption{Downstream evaluation on Passkey Retrieval and Qasper. \model enables Llama-2 to consistently outperform both the original model and the baseline that truncates inputs to 4K. The truncation baseline drops excessive tokens altogether when the context is longer than the model's pretraining length, keeping only the most recent ones, which happens before the forward pass starts without changing the attention mechanism. 
}
\label{tab:downstream}
\end{table*}

%% file: tables/generation.tex
\begin{table*}[t!]
\centering
\linespread{1}


\begin{adjustbox}{max width=\textwidth}
\begin{tabular}{@{} l cccccc m{0.01em} ccccc @{}}
\toprule

& \multicolumn{5}{c}{\textbf{BLEU}} 
&& \multicolumn{5}{c}{\textbf{ROUGE}}\\
\cline{3-7}
\cline{9-13}
\textbf{Model} & $L_\text{pretrain}$ & \phantom{0}2K & \phantom{0}4K & \phantom{0}8K & 16K & 32K
&& \phantom{0}2K & \phantom{0}4K & \phantom{0}8K & 16K & 32K  \\
\midrule
\multicolumn{12}{@{}l}{\textit{ArXiv}}\\
MPT-7B & 4K & \phantom{0}0.0 & \phantom{0}0.2 & \phantom{0}0.0 & \phantom{0}0.0 & \phantom{0}0.4 && \phantom{0}5.6 & \phantom{0}3.6 & \phantom{0}5.9 & \phantom{0}1.7 & \phantom{0}1.4\\
MPT-7B-SW & 65K & 16.6 & 21.5 & 15.2 & 18.9 & 14.8 && 26.5 & 30.1 & 26.6 & 27.4 & \textbf{27.0}\\
MPT-7B + \model{} & 4K & 16.1 & 20.2 & 12.6 & 13.9 & \textbf{19.7} && 23.8 & 24.9 & 24.1 & 29.0 & 26.6\\
Llama-2 & 4K & 26.6 & \phantom{0}0.0 & \phantom{0}0.0 & \phantom{0}0.0 & \phantom{0}0.0 && 31.4 & \phantom{0}0.2 & \phantom{0}0.0 & \phantom{0}0.0 & \phantom{0}0.0\\
Llama-2 + \model{} & 4K & \textbf{26.9} & \textbf{23.6} & \textbf{23.9} & \textbf{24.8} & 18.4 && \textbf{31.8} & \textbf{30.9} & \textbf{28.8} & \textbf{29.2} & 20.4\\
\midrule

\multicolumn{12}{@{}l}{\textit{OpenWebText2}}\\
MPT-7B & 4K & \phantom{0}0.9 & \phantom{0}0.9 & \phantom{0}1.0 & \phantom{0}1.0 & - && \phantom{0}7.5 & \phantom{0}6.6 & \phantom{0}6.4 & \phantom{0}6.8 & -\\
MPT-7B-SW & 65K & \phantom{0}8.4 & \phantom{0}6.1 & \phantom{0}7.5 & \phantom{0}8.4 & - && 21.0 & 19.3 & 18.5 & \textbf{22.0} & -\\
MPT-7B + \model{} & 4K & 5.0 & \phantom{0}4.1 & \phantom{0}5.1 & \phantom{0}2.8 & - && 16.6 & 15.4 &16.2 & 16.0 & -\\
Llama-2 & 4K & \phantom{0}8.8 & \phantom{0}0.0 & \phantom{0}0.0 & \phantom{0}0.0 & - && \textbf{22.4} & \phantom{0}0.2 & \phantom{0}0.0 & \phantom{0}0.0 & -\\
Llama-2 + \model{} & 4K & \phantom{0}\textbf{9.0} & \phantom{0}\textbf{7.2} & \phantom{0}\textbf{9.7} &\phantom{0}\textbf{9.6} & - && 21.9 & \textbf{21.2} & \textbf{19.6} & 19.6 & -\\
\bottomrule

\end{tabular}
\end{adjustbox}
\caption{Text generation quality on ArXiv and OpenWebText2. \model{} consistently generalizes the generation quality to extreme lengths, achieving performance that is comparable to or better than the fine-tuned LLM, MPT-7B-Storywriter. Some 0 BLEU scores are caused by the poor generation quality from the vanilla LLMs as they generate mostly nonsensical texts.
}

\vspace{-4mm}
\label{tab:generation}
\end{table*}

%% file: body/5_2_ablation.tex
\subsection{Ablation Study}
\label{sec:ablation}
Figure~\ref{fig:ablation} provides an ablation study with the LLaMA model on the ArXiv dataset about why both components in \model{} are essential for maintaining LLM functionality over the length of 8K. We compare \model with its variants to show the efficacy of the design and also to validate the factors in \S\ref{sec:diagnosis}.
Among all curves, only \model has relatively stable log-perplexity, meaning that components in \model are all essential for successful length generalization.
The vanilla LLM model (the ``vanilla'' curve) fails immediately with exploding NLL.
If we only apply $\mathbf{\Lambda}$-shaped mask (the ``$\Lambda$'' curve) and do not bound inter-token distance (Factor 1), the NLL still explodes immediately after pre-training lengths.
The ``ceiling'' curve only applies the distance ceiling technique but not the $\Lambda$-shaped mask to limit the number of attended tokens. The performance still degenerates (evidenced by an ever-increasing NLL). This confirms that the existence of Factor 2, too many tokens, is still harming the LLMs' performance.
The ``window'' curve shows a baseline with the sliding-window attention pattern, which only attends to the most recent tokens in a sliding window without altering the input text. It produces the second worst NLL values, which indicates a significant performance and fluency degradation. This confirms our theoretical analysis of factor 3. Due to its visibly much worse performance, we exclude it from other evaluations.

Another similar baseline to ``window'' is the truncation baseline, which drops excessive tokens altogether when the context is longer than the model’s pre-training length, keeping only the most recent ones. This truncation process happens before the forward pass starts and essentially removes the truncated text from the input to the model without changing the attention mechanism. We compared this baseline in two places in the paper. In \S\ref{sec:downstream} and Table~\ref{tab:downstream}, LM-infinite outperforms this baseline on downstream tasks. In Section~\ref{sec:generation} and Figure~\ref{fig:truncation_comparison}, LM-infinite achieves a better trade-off between computation complexity and generation quality than this baseline.

%% file: body/5_3_downstream.tex
\subsection{Downstream Evaluation}
\label{sec:downstream}
As LLMs are often deployed for downstream tasks, we evaluate how \model{} performs on two long-input tasks under the zero-shot setting: Passkey Retrieval~\citep{mohtashami2023landmark} and Qapser~\cite{dasigi2021dataset}. Passkey Retrieval buries a passkey at a random position in a long distraction text and, in the end, asks what the passkey is. Qasper is a question-answering dataset on scientific papers with a total of 1.5K testing question-answer pairs. 
We evaluate Llama-2-7b-chat, as its instruction tuning enables good task-solving ability~\citep{bai2023longbench}, with middle top-5 tokens enabled on higher than 5-th layer (see \S\ref{sec:approach} for definition and Appendix~\ref{sec:evaluation_details} for hyperparameter selection). 
Results are listed in Table~\ref{tab:downstream}. \model{} consistently outperforms the baselines on both tasks, with a 37.2 percentage gain on Passkey Retrieval and a 1.2 percentage gain on the Qasper task.
Passkey retrieval locates useful information uniformly in a sequence, so the performance of the truncated baseline largely depends on whether the remaining part covers the passkey. On Qasper, the top-$k$ attention is necessary for achieving good performance, which indicates that similarly important information in the middle needs to be attended to.
This suggests that it can improve downstream task performance on long inputs without fine-tuning while the vanilla model immediately fails.

%% file: figures/truncation_comparison.tex

\begin{figure}
\centering
\includegraphics[width=0.48\textwidth]{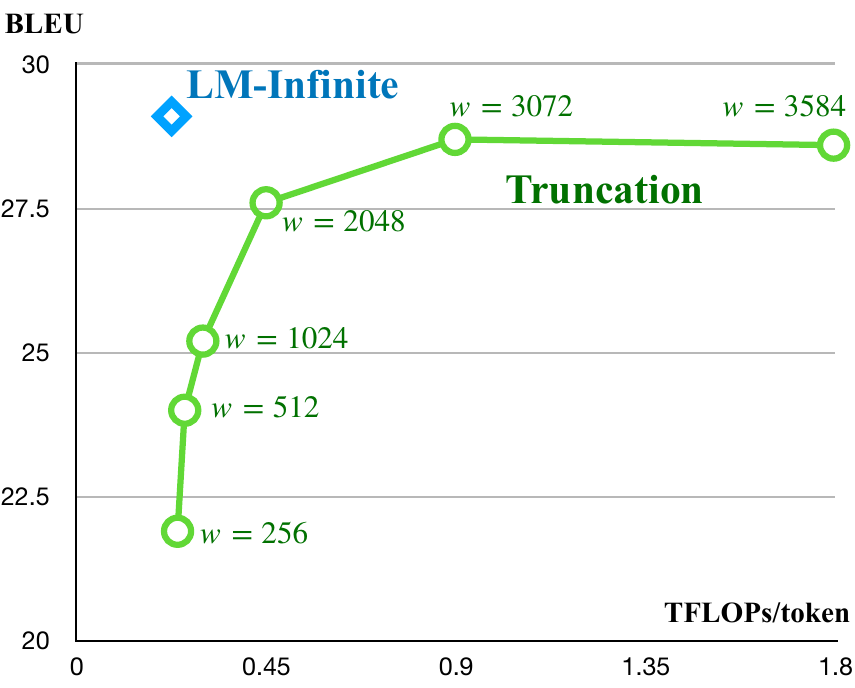}

\caption{\model{} achieves a better trade-off between computation complexity with generation quality than simple truncation.}
\label{fig:truncation_comparison}
\end{figure}

%% file: body/5_4_generation.tex
\subsection{Generation Quality}
\label{sec:generation}
We further evaluate \model{}'s generation quality on ArXiv and OpenWebText2 test sets, with BLEU~\citep{papineni2002bleu} and ROUGE~\citep{lin2004rouge} (ROUGE-L).
We let the LLMs generate 100 tokens after each milestone length and use the following 100 tokens in original texts as references. As the generation is time-consuming, we sample 100 long sequences for evaluation for each dataset.
The results are summarized in Table~\ref{tab:generation}. The trend is similar to that in the last section: \emph{without} parameter updates, \model{} successfully allows LLMs to retain their generation quality while generating sequences longer than training, comparable to the fine-tuned baselines such as MPT-7B-SW.
The generation results from the vanilla LLMs are poor and contain mostly nonsensical texts, resulting in many close-to-zero scores. For some BLEU scores, it yields zero \{2,3,4\}-gram overlaps with the reference texts. As BLEU is weighted geometric mean over \{1,2,3,4\}-gram precisions, the final BLEU scores for those columns are 0. Appendix Table~\ref{tab:example} presents some generation output examples that can provide a good picture of the generation quality.
We also evaluate the efficiency in Appendix~\ref{sec:efficiency}: with 32K-long sequences, \model{} achieves 2.7$\times$ decoding speedup and 7.5$\times$ GPU memory saving.

A few example generations are shown in Appendix~\ref{sec:example}. 
We also compare \model{} with a simple truncation-based baseline by repeatedly truncating excessive contexts. However, as the generation length increases, frequent truncations and re-encoding of new contexts are required. The larger the truncation window is, the more context is kept, but the larger the computational overhead. 
We let the models generate 10k tokens on ArXiv. In Figure~\ref{fig:truncation_comparison}, it is clear that \model achieves a substantially better quality-efficiency tradeoff. With similar computation, \model{} outperforms the baseline by about 5 BLEU.
To achieve a similar BLEU, \model{} incurs only $<$25\% computational overhead than the truncation baseline.

%% file: body/6_conclusion.tex
\section{Conclusions and Future Work}
\label{sec:conclusion}

This work proposes a zero-shot length generalization method for various off-the-shelf LLMs without parameter updates. Through theoretical analysis and empirical investigation, this work identifies three major factors contributing to this length generalization failure. Our theoretical analysis further reveals why truncating the attention window and relative positional encodings are inadequate to address them. Our solution, \model, is a simple and effective method for enhancing LLMs' capabilities of handling long contexts. It allows LLMs pre-trained with 2K or 4K-long segments to generalize to up to 200M length inputs while retaining perplexity.
It also improves performance on downstream tasks such as Passkey Retrieval and Qasper in the zero-shot setting. It brings substantial efficiency improvements: a 2.7$\times$ decoding speed up and a 7.5$\times$ memory saving over the original model. \model's computational efficiency and ease of use allow researchers without enormous computational resources to use LLMs on long sequences. 
Future work can investigate if these techniques allow for more efficient and effective LLM pre-training and fine-tuning. Another direction is to apply \model{} to applications such as long reasoning, long-dialogue, retrieval-augmented generation, or long literature generation.

%% file: body/7_limitations.tex
\section*{Limitations}
This work evaluates a wide range of open-domain LLMs. However, without access to the source code of proprietary LLMs such as ChatGPT, the proposed method could not be evaluated on them. Furthermore, due to limited computational resources and time, the proposed method has not been evaluated on texts with even larger lengths, such as 1G. The model is designed on relative positional encoding Transformer models, which is the mainstream backbone for most modern LLMs. The question of how \model can enable more efficient fine-tuning or pre-training can also be explored in future work.

%% file: body/8_acknowledgement.tex
\subsection*{Acknowledgement}

This research is partially supported by U.S. DARPA KAIROS Program No. FA8750-19-2-1004, and DARPA INCAS Program No. HR001121C0165. The views and conclusions contained herein are those of the authors and should not be interpreted as necessarily representing the official policies, either expressed or implied, of DARPA, or the U.S. Government. The U.S. Government is authorized to reproduce and distribute reprints for governmental purposes, notwithstanding any copyright annotation therein.

%% file: body/A_implementation.tex
\section{Implementation Details}
\label{sec:evaluation_details}

In this section, we introduce some implementation details of \model as well as the hyper-parameter selection.

\subsection{Computational resources}

All experiments are run on single A100 GPUs with 80GB GPU memory each. The 200M length generalization runs for 20 hours. The downstream tasks take 3$\sim$7 hours to evaluate each. Our work does not involve any training or fine-tuning. All models are loaded with Huggingface\footnote{\url{https://huggingface.co}https://huggingface.co} code repository. Rouge and BLEU scores are loaded from \texttt{evaluate}\footnote{\url{https://huggingface.co/docs/evaluate/index}} package. Datasets and models are used with permission from their licenses.

\subsection{The size of starting attention span}

We vary the value of $n_\text{starting}$ and find \model to be tolerant with it taking a wide range of values without sacrificing the NLL values. Speicifically, we evaluate it on sequences of 16k length in the ArXiv validation set and calculate the average NLL.

\begin{table}[ht]
\centering
\linespread{1}
\begin{tabular}{ccccccc}
\toprule
\multicolumn{7}{c}{$n_\text{starting}$} \\
0 & 1 & 2 & 10 & 100 & 1000 & 2000 \\
\midrule
6.43 & 1.03 & 1.03 & 1.02 & 1.02 & 1.81 & 4.96 \\
\bottomrule
\end{tabular}
\caption{Effect on \model's NLL by varying $n_\text{starting}$.}
\end{table}

\subsection{Reintroducing Top-$k$ Middle Tokens}

This optional technique involves optionally attending to $k$ tokens in the middle with the largest attention logits.
Here, the $k$ tokens are selected independently for each attention head and only apply to layers higher than $h$-th layer. These tokens have an attention distance of $d=\frac{1}{2}L_\text{pre-train}$. We select these hyper-parameters based on a held-out validation set of Passkey Retrieval. On Llama-2, we use $k=5$ and $h=5$. As an ablation study, we vary each hyper-parameter and observe its effects on Passkey Retrieval accuracy.

\begin{table}[h]
\centering
\linespread{1}
\begin{tabular}{ccccccc}
\toprule
\multicolumn{7}{c}{$k$} \\
1 & 3 & 5 & 10 & 20 & 50 & 200 \\
\midrule
0.69 & 0.81 & 0.81 & 0.79 & 0.8 & 0.79 & 0.73 \\
\bottomrule
\end{tabular}
\caption{Effect of varying $k$.}
\end{table}

\begin{table}[h]
\centering
\linespread{1}
\begin{tabular}{ccccc}
\toprule
\multicolumn{5}{c}{\textbf{attention distance}} \\
512 & 1024 & 2048 & 3072 & 4096 \\
\midrule
0.78 & 0.79 & 0.81 & 0.68 & 0.63 \\
\bottomrule
\end{tabular}
\caption{Effect of attention distance of the middle tokens.}
\end{table}

\begin{table}[ht]
\centering
\linespread{1}
\begin{tabular}{ccccccc}
\toprule
\multicolumn{7}{c}{$h$} \\
0 & 4 & 5 & 6 & 8 & 16 & 24 \\
\midrule
0.81 & 0.94 & 0.94 & 0.91 & 0.46 & 0.45 & 0.46 \\
\bottomrule
\end{tabular}
\caption{Effect of varying $h$.}
\end{table}

On the Qasper dataset, for both vanilla models and \model, we use 6K sub-sequence of inputs as prompts and use a systematic prompt format described in Llama-2 paper~\citep{touvron2023llama2}.

%% file: body/A_related_work.tex
\section{Additional Related Work}
\label{sec:more_related}

After our preprint, there have been papers that cite our work and investigate zero-shot or few-shot length generalization of LLMs. 
As many absolute or relative position encoding methods are based on periodic functions, \cite{qu2023gpt, ding2024longrope, liu2023scaling, jiang2023longllmlingua} propose to apply LLMs (fine-tuned or not) on decreased period frequencies (which is equivalent to interpolating position indices) to adapt LLMs to longer sequences.
Some other papers finetune LLMs with designed attention patterns~\cite{oren2024transformers, zhang2024soaring} on long contexts, using neural tangent kernel~\citep{peng2023yarn}, or with low-rank adaptation(LoRA)~\citep{chen2023longlora}.
\cite{yang2024attendre} instead proposes a wait-to-attend mechanism to extend length limits for memory-based Transformers.
Other ways of key-value cache selection/eviction methods are investigated in \cite{ren2024efficacy, dong2024get, zhang2024h2o}.
Similarly, \cite{huang2023boosting, lee2024human} tackles long context by learning to dynamically prune, select, or summarize contexts.
Alternatively, context compression methods~\citep{shao2024flexibly} propose to learn to compress long contexts into shorter embedding sequences.
Some work proposes alternative position encodings~\citep{song2023zebra, zhang2024found, zhu2023pose} or landmark token embeddings~\citep{luo2024bge} that enable extendable context limits.
~\cite{xiao2024efficient} is a later concurrent work to ours with a similar approach to LLM length generalization. Unlike our work, they feed a sequence to an LLM token-by-token which limits their extreme length generalization (4M v.s. 200M of ours), and more importantly, they do not show improvements on downstream tasks without pre-training an LLM from scratch.
Finally, there is a lot of new benchmarks~\cite {qiu2024clongeval, zhang2024infty, yuan2024lv, wang2024novelqa, lv2024longwanjuan, bai2024longalign} proposed to evaluate the long-context performance of LLMs.

%% file: body/A_proof1.tex
\section{Formal Statement of Theorem~\ref{thm:attention_logit_explosion_informal}}
\label{sec:proof_explosion}

Let us denote the logit function with relative position encoding as $w(\mathbf{q}, \mathbf{k}, d)\in\mathbb{R}$. It maps the query $\mathbf{q}$, key $\mathbf{k}$, and their distance $d$, to an attention logit. 
The final attention weights are usually calculated by a softmax operation. For example, given $n$ tokens with indices $(1, \cdots, n)$, the attention by the last token on a preceding token at position $i$ is:
\begin{equation}
    \text{Attn}(\text{token}_n, \text{token}_i) = 
    \frac{e^{w(\mathbf{q}_n, \mathbf{k}_i, n-i)}}{\sum_{j=1}^{n}e^{w(\mathbf{q}_n, \mathbf{k}_j, n-j)}}
\end{equation}

Then the formal theorem of Theorem~\ref{thm:attention_logit_explosion_informal} is as follows:
\begin{theorem}
\label{thm:attention_logit_explosion}
(Formal)
    Let $\mathbf{q}$ and $\mathbf{k}$ be random vectors sampled from training distributions $\mathcal{D}_\mathbf{q}$ and $\mathcal{D}_\mathbf{k}$, respectively, where $\mathcal{D}_\mathbf{q}$ and $\mathcal{D}_\mathbf{k}$ are the trained distributions for $\mathbf{q}$ and $\mathbf{k}$, respectively.
    We use the pseudo-dimension $\dim_P(\cdot)$ defined in \cite{pollard1990empirical}, which measures the representation capacity of a function family. Assume that the set of distance-based logit functions $\mathcal{H} = \{w(\cdot, \cdot, d) | d \in \mathbb{N}\}$ has bounded pseudo-dimension $\dim_P(\mathcal{H}) = r$\footnote{This is true for most current techniques. See discussions in Appendix~\ref{sec:pseudo_dimension}}.
    Let us also define the distinguish-ability of two distances $d$ and $d'$ under $w$ as their expected squared difference:
    $\mu_w(d, d') = \mathbb{E}_{\mathbf{q}\sim\mathcal{D}_\mathbf{q},\mathbf{k}\sim\mathcal{D}_\mathbf{k}}(w(\mathbf{q}, \mathbf{k}, d) - w(\mathbf{q}, \mathbf{k}, d'))^2$. 
    We assume that $w$ is not limited to recognizing only a finite group of distances, otherwise, all distances longer than a threshold will become almost the same as shorter distances. Formally, for any $n$, there is a partition of $[0 .. n ]$ into $\alpha(n)$ groups so that, $\mu_w( d, d')\leq\epsilon$ for any $d, d'$ from the same group. $\alpha(n)\in\mathbb{N}$ is non-decreasing and unbounded function. Then we have:
    \[
        \sup_{\mathbf{q}, \mathbf{k}, d\leq n} |w(\mathbf{q}, \mathbf{k}, d)| \geq \left(\frac{\alpha(n)}{2}\right)^\frac{1}{2r}\frac{\epsilon}{4e}.
    \]
\end{theorem}

We first borrow a lemma from ~\cite{haussler2018decision}, which we paste below.  Note that a cover size $\mathcal{N}(\epsilon, \mathcal{H}, \mu)$ is defined as the minimum cardinal of a cover-set $\mathcal{H}'$ so that any element of $h\in\mathcal{H}$ will be within $\epsilon$ distance to at least one element $h' \in \mathcal{H}'$.

\begin{lemma}
    Let $\mathcal{H}$ be a function family mapping from space $X$ to range $[0, B]$, and its pseudo-dimension $\dim_P (\mathcal{H})=r$. Then for any probabilistic measure $P$ on $X$, and $\epsilon \in [0, B]$, we have that the $\epsilon$ cover of $\mathcal{H}$ under metric $\mu(h_1, h_2)=\mathbb{E}_{x\sim P}(h_1(x) - h_2(x))^2$ is bounded by:
    \[
    \mathcal{N}_P (\epsilon, \mathcal{H}, \mu) \leq 2\left(\frac{2eB}{\epsilon} \ln \frac{2eB}{\epsilon}\right)^r
    \]
\end{lemma}

With this lemma, we can go on to prove Theorem~\ref{thm:attention_logit_explosion} as follows.

\begin{proof}
    We prove by contradiction.
Assume that $\sup_{\mathbf{q}, \mathbf{k}, d\leq n} |w(\mathbf{q}, \mathbf{k}, d)| < a = \left(\frac{\alpha(n)}{2}\right)^\frac{1}{2r}\frac{\epsilon}{4e}$. Without loss of generality, we can shift all the values to range $[0, 2a]$. Then the function family $\mathcal{H} = \{w(\cdot, \cdot, d) | d \in \mathbb{N}\}$  will have cover size $\mathcal{N}_P (\epsilon, \mathcal{H}, \mu) \leq 2\left(\frac{4ea}{\epsilon} \ln \frac{4ea}{\epsilon}\right)^r < \alpha(n)$.

However, this is smaller than the number of different distances and relative weight attentions $\mathcal{H}$, which means that at least 2 functions will be close to each other $(w(\cdot, \cdot, d), w(\cdot, \cdot, d'))^2 < \epsilon$. This constitutes a contradiction with the distinguishability condition.

\end{proof}

%% file: body/A_proof2.tex
\section{Formal Statement and Proof of Proposition~\ref{thm:entropy_explosion_informal}}
\label{sec:entropy_explosion}

The formal statement of Proposition~\ref{thm:entropy_explosion_informal} is the following:

\begin{proposition} \textbf{(Attention Entropy Explosion)}\label{thm:entropy_explosion}
    Let $w_1, w_2, \cdots, w_n \in [-B, B]$ be a sequence of attention logits. Then the entropy of the attention distribution they span is asymptotically lower bounded by $\ln n$: 
    \[
        H\left(\left(\frac{e^{w_i}}{\sum_{j=1}^{n}e^{w_j}} \bigl\vert 1 \leq i \leq n \right)\right) = \Omega(\ln n)
    \]
\end{proposition}
\noindent The entropy approaches $+\infty$ as $n$ grows larger.

\begin{proof}
    Note that entropy on a discrete distribution is defined as $\text{Entropy}(P) = - \sum_i p_i \ln p_i$. Then the attention entropy determined by attention logits $\{w_i|1\leq i \leq n\}$ is 
    \begin{align*}
        & \text{Entropy} (\text{Attention}) \\
        =& - \sum_i \frac{e^{w_i}}{\sum_j e^{w_j}} \ln \frac{e^{w_i}}{\sum_j e^{w_j}} \\
        =& - \sum_i \frac{e^{w_i}}{\sum_j e^{w_j}} \left( w_i - \ln \sum_j e^{w_j} \right) \\
        =& - \sum_i \frac{e^{w_i}}{\sum_j e^{w_j}}  w_i + \ln \sum_j e^{w_j} \\
        \geq& -\max_i w_i + \ln (n e^{-B}) \\
        \geq& \ln n - 2B \\
        =& \Omega(\ln n )
    \end{align*}
\end{proof}

%% file: body/A_layer_positions.tex
\section{More on Implicitly Encoded Positions}
\label{sec:layer_positions}

\input{figures/layer_positions}

We also plot the token features of more layers with PCA projection to the 2D plane in Figure~\ref{fig:layer_positions}. We see that from layer 2 to higher layers, the initial few tokens occupy a distinct region with later tokens. Therefore, if these tokens are discarded by window attention during attention, the attention output, which is a weighted sum of $v_i$ vectors, will reside in a different region. This can explain why windowed attention does not work and why the first few tokens need to be kept by our  $\mathbf{\Lambda}$-shaped attention.

%% file: figures/layer_positions.tex
\begin{figure*}[t]
\centering
\includegraphics[width=\textwidth]{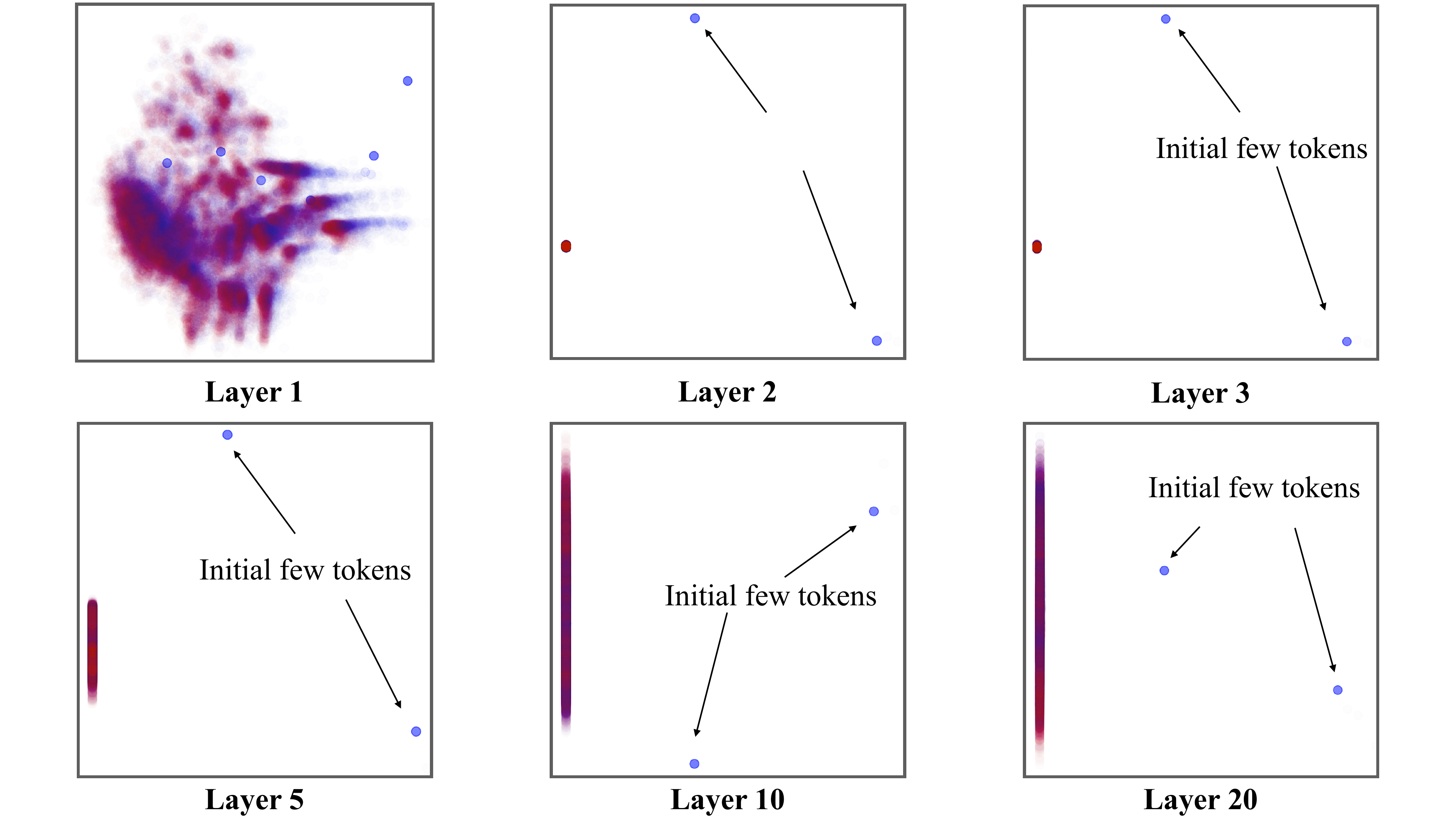}

\caption{In Llama, at second or higher layers, the initial few tokens encode a strong position signal and occupy a distinct feature region. Abandoning them might move the attention output vector out of the pre-training distribution.}
\label{fig:layer_positions}
\end{figure*}

%% file: body/A_pseudo_dimension.tex
\section{Pseudo-Dimension Assumption on Attention Logit Functions}
\label{sec:pseudo_dimension}

In Theorem~\ref{thm:attention_logit_explosion}, we assumed that the family of distance-based logit functions $\mathcal{H}=\{w(\cdot, \cdot, d) | d\in\mathbb{N}\}$ has a finite pseud-dimension. In this section, we demonstrate that most current implementations of relative positional encodings do have a finite pseudo-dimension. Let us discuss a few examples in the following:

\paragraph{T5-Bias and Alibi}
It is easy to see that, the difference between any two logit functions is uniform:
$w(\cdot, \cdot, d_1) - w(\cdot, \cdot, d_2) = \text{bias}(d_1) - \text{bias}(d_2)$ regardless of the input. Therefore this family cannot shatter any set larger than 2, so the pseudo-dimension is 1.

\paragraph{Windowed Attention}
This operation is equivalent to limiting the family to a finite size $|\mathcal{H}| = d_{\max{}}+1$, where $d_{\max{}}$ is the size of the window. Therefore $\dim_P ( \mathcal{H} ) \leq d_{\max{}}+1$.

\paragraph{NoPE}
As there is no explicit positional encoding implemented, all distances are equivalent. The pseudo-dimension is 1.

\paragraph{RoPE, CAPE, and XPos}
For RoPE, the logit function $w(\mathbf{q}, \mathbf{k}, d)$ is the weighted sum of finite fixed sinusoidal functions $\{\sin(\omega_i d), \cos(\omega_i d)\}$. The size of this set is equivalent to the feature dimension number $k$. We know that $\dim_P(\mathcal{H}_1 + \mathcal{H}_1) \leq \dim_P(\mathcal{H}_1) + \dim_P(\mathcal{H}_2)$. Also, the scaling of a single function can only have a pseudo-dimension of 2. Therefore, the whole family has a bounded pseudo-dimension $\dim_P(\mathcal{H}) \leq 2k$. The analysis on CAPE and XPos is similar.

%% file: body/A_efficiency.tex
\section{Computational Efficiency Evaluation}
\label{sec:efficiency}
To evaluate the computational efficiency of \model{}, we run the Llama-2-7B model on 100 sequences of 32k length in the ArXiv dataset. The hardware is a single A100 GPU with 80GB GPU memory. As the memory is not enough to host the whole computation graph during decoding, we use DeepSpeed~\citep{rasley2020deepspeed} with Zero3 optimization. We also have to modify the computation code to further reduce GPU memory usage and prevent out-of-memory errors. With that in mind, the vanilla Llama-2-7B model encodes with an average speed of 48.19s per sequence, while \model{} encodes with an average of 15.26s per sequence, a 3.16x speedup. The vanilla Llama-2-7B model decodes with 7.34s per token, while \model{} decodes with 2.70s per token, a 2.72x speedup.
We also evaluate the GPU memory usage on 32k-length inputs, the statistics of which are profiled with PyTorch Profiler. The vanilla model uses 33.2Gb memory per sequence, while \model uses 4.41Gb per sequence, a 7.53$\times$ memory saving.

%% file: body/A_example_generation.tex
\section{Example Generation}
\label{sec:example}

\begin{table*}[t!]
\centering
\small
\linespread{1}


\setlength{\tabcolsep}{2mm}{

\caption{Example text generations on ArXiv and OpenWebText2 corpora after 8k context lengths.}
\begin{tabular}{lp{110mm}}
\toprule

 & \textbf{ArXiv Dataset} \\

\midrule

\textbf{Context}
& \textit{Our proof of the Kottwitz formula imitates these arguments of Langlands and Rapoport of deriving the Kottwitz formula from their conjecture. As such, their theory of Galois gerbs is a major ingredient in this work. Another essential ingredient is a recent work of Kisin [@Kisin17] on the aforementioned conjecture of Langlands and Rapoport. Previously, Kottwitz [@Kottwitz92] proved the formula in PEL-type cases (of simple Lie type $A$ or $C$) by a method which is based on the Honda-Tate theory. This method however cannot be applied in general Hodge-type situations, and indeed} \\

\midrule

\textbf{Vanilla Llama-2}
& \textit{of this (of over-equary-her, and [( and, in the...cister '-- and an of the model to by. by ..., this, by the. It, and it, 7. --(of an equist (of the. to the [[[WNE (to. and for the (((de in the (for the andistile–c.-[de (for in an inc ort, ort (betness in >with (with, based (and (>~such ((c of a or for the abstract as. of *. (up. W in the, of the or an. and of a.0 in (and of (over in such [near (---of (([her (pro -- [p for [with the of a: their in the in of '[alism in the, [(du of that of the ((both [equ of the of [for ([of as of ... and of the of their of ((---for of the of the (t of the of the of the of his of the. of the. of the of it (D with the d. and of of the of the of the of [of of a of a, [[for for a per of a of its of a (for of the and of the} \\

\midrule

\textbf{\model{}}
& \textit{Kottwitz [@Kottwitz84a] had formulated the conjecture, but at the time no idea had yet been established on how to prove it. This conjecture involves some algebraic statements of automorphic forms which have an algebraic nature, but are only valid for automorphic forms (which are not holomorphic automorphic forms [@Kaplansky1966]), whose (abstract) objects are Galois extensions of $F$ and $\bar{F}$ (which in general are not necessarily finite extensions), in other words a $p$-valued Hodge-type structure (in $n$-dimension). The Langlands-Rapoport conjecture is now established [@ScholzeShin13] whose formulation is in terms of (a more explicit class of) Hodge-type structures rather than these abstract Galois objects, and the automorphic side for $p$-adic $L$-functions are due to Kisin [@Kisin15].} \\

\midrule

& \textbf{OpenWebText2} \\

\midrule

\textbf{Context}
& \textit{This auto-structuration of a complete whole is defined as "auto-regulation". In the individual, the latter is established by biological rhythms, biological \& mental regulations and mental operations. These can be theoretically formalized.
Piaget refuses to accept that "real" dialectical tensions between physical objects are the true foundations of thought and cognition (its possibility, genesis \& progressive development). Piaget never fills in what reality is. He maintains no ontological view on reality-as-such, considered to be the borderline of both the developing subject and its objective world, stage after stage.
The cognitive is approached as a process, for rationality grows in developmental stages, each calling for a particular cognitive structure on the side of} \\

\midrule

\textbf{Vanilla Llama-2}
& \textit{the in that, of it and and \', " of the. of " of " of the pr of of  of the of the of the [. of “ – s (1 ( of s of the " of quends of it \&nd of trail of the, the \&, 1. – a and at-and and and the " for on the is, core by " ( as (:: a. -. ( on its to “ is a, ( on, th is of all but". ".. in the, in the the the and in the for the. , and of \' of the or of the :! of " A. of the of the, of the of the of [of the of the: of a th, and s of the of the the as of the, B of h. the —. the : a, and, in a: \& (se and [ ( (t, d of the and....:,. of the (,. ( A. A. Pro: T. I " (b in the \' of the, of  of: of " of, of the of the [ of, (: - " of the , of s of all ma of the [ and of the of the of [ of the of the of the ( the \& ( w [ (:, and ad and and [ " of a in. for the, of \' in (} \\

\midrule

\textbf{\model{}}
& \textit{cognition. It develops gradually. For Piaget, the cognitive originates in experience as it is subjected to structure : i.e. as it is structured, ordered and transformed. In the first stage, the "sensorial stage", structures are formed and integrated based on perception. These structures are built by organic, biological and psychic transformations, which lead to experience. The "concrete operational stage" of intellectual development refers to a stage when the mind of the child starts to use the logic of the general rule and the rule of the general case. There is now a logical, conceptual and operational distinction of concepts. Reasoning is made explicit by applying logical operations, such as subtraction, addition and multiplication to the "mental" object as well as to the "perceived" world of reality. The child\'s logic can now make use of the logical operations, though for him (her) it is a conceptual understanding of abstract operations. This is Piaget\'s concept of "genetic development". In the "formal operational stage", logical operations are combined using logical or conceptual structures.} \\

\bottomrule
\end{tabular}
\label{tab:example}
}
\end{table*}

%% file: main.bbl
\begin{thebibliography}{73}
\expandafter\ifx\csname natexlab\endcsname\relax\def\natexlab#1{#1}\fi

\bibitem[{Anil et~al.(2022)Anil, Wu, Andreassen, Lewkowycz, Misra, Ramasesh,
  Slone, Gur-Ari, Dyer, and Neyshabur}]{anil2022exploring}
Cem Anil, Yuhuai Wu, Anders Andreassen, Aitor Lewkowycz, Vedant Misra, Vinay
  Ramasesh, Ambrose Slone, Guy Gur-Ari, Ethan Dyer, and Behnam Neyshabur. 2022.
\newblock Exploring length generalization in large language models.
\newblock \emph{Advances in Neural Information Processing Systems},
  35:38546--38556.

\bibitem[{Bai et~al.(2024)Bai, Lv, Zhang, He, Qi, Hou, Tang, Dong, and
  Li}]{bai2024longalign}
Yushi Bai, Xin Lv, Jiajie Zhang, Yuze He, Ji~Qi, Lei Hou, Jie Tang, Yuxiao
  Dong, and Juanzi Li. 2024.
\newblock Longalign: A recipe for long context alignment of large language
  models.
\newblock \emph{arXiv preprint arXiv:2401.18058}.

\bibitem[{Bai et~al.(2023)Bai, Lv, Zhang, Lyu, Tang, Huang, Du, Liu, Zeng, Hou,
  Dong, Tang, and Li}]{bai2023longbench}
Yushi Bai, Xin Lv, Jiajie Zhang, Hongchang Lyu, Jiankai Tang, Zhidian Huang,
  Zhengxiao Du, Xiao Liu, Aohan Zeng, Lei Hou, Yuxiao Dong, Jie Tang, and
  Juanzi Li. 2023.
\newblock \href {http://arxiv.org/abs/2308.14508} {Longbench: A bilingual,
  multitask benchmark for long context understanding}.

\bibitem[{Beltagy et~al.(2020)Beltagy, Peters, and
  Cohan}]{beltagy2020longformer}
Iz~Beltagy, Matthew~E Peters, and Arman Cohan. 2020.
\newblock Longformer: The long-document transformer.
\newblock \emph{arXiv preprint arXiv:2004.05150}.

\bibitem[{Borgeaud et~al.(2022)Borgeaud, Mensch, Hoffmann, Cai, Rutherford,
  Millican, Van Den~Driessche, Lespiau, Damoc, Clark
  et~al.}]{borgeaud2022improving}
Sebastian Borgeaud, Arthur Mensch, Jordan Hoffmann, Trevor Cai, Eliza
  Rutherford, Katie Millican, George~Bm Van Den~Driessche, Jean-Baptiste
  Lespiau, Bogdan Damoc, Aidan Clark, et~al. 2022.
\newblock Improving language models by retrieving from trillions of tokens.
\newblock In \emph{International conference on machine learning}, pages
  2206--2240. PMLR.

\bibitem[{Bueno et~al.(2022)Bueno, Gemmell, Dalton, Lotufo, and
  Nogueira}]{bueno2022induced}
Mirelle~Candida Bueno, Carlos Gemmell, Jeff Dalton, Roberto Lotufo, and Rodrigo
  Nogueira. 2022.
\newblock Induced natural language rationales and interleaved markup tokens
  enable extrapolation in large language models.
\newblock In \emph{Proceedings of the 1st Workshop on Mathematical Natural
  Language Processing (MathNLP)}, pages 17--24.

\bibitem[{Chen et~al.(2023{\natexlab{a}})Chen, Wong, Chen, and
  Tian}]{chen2023extending}
Shouyuan Chen, Sherman Wong, Liangjian Chen, and Yuandong Tian.
  2023{\natexlab{a}}.
\newblock Extending context window of large language models via positional
  interpolation.
\newblock \emph{arXiv preprint arXiv:2306.15595}.

\bibitem[{Chen et~al.(2021)Chen, Zeng, Ji, and Yang}]{skyformer2021}
Yifan Chen, Qi~Zeng, Heng Ji, and Yun Yang. 2021.
\newblock Skyformer: Remodel self-attention with gaussian kernel and nyström
  method.
\newblock In \emph{Proc. Thirty-fifth Annual Conference on Neural Information
  Processing Systems (NeurIPS2021)}.

\bibitem[{Chen et~al.(2022)Chen, Zeng, Ji, and Yang}]{Sketching2022}
Yifan Chen, Qi~Zeng, Heng Ji, and Yun Yang. 2022.
\newblock Sketching as a tool for understanding and accelerating self-attention
  for long sequences.
\newblock In \emph{Proc. The 2022 Conference of the North American Chapter of
  the Association for Computational Linguistics - Human Language Technologies
  (NAACL-HLT2022)}.

\bibitem[{Chen et~al.(2023{\natexlab{b}})Chen, Qian, Tang, Lai, Liu, Han, and
  Jia}]{chen2023longlora}
Yukang Chen, Shengju Qian, Haotian Tang, Xin Lai, Zhijian Liu, Song Han, and
  Jiaya Jia. 2023{\natexlab{b}}.
\newblock Longlora: Efficient fine-tuning of long-context large language
  models.
\newblock In \emph{The Twelfth International Conference on Learning
  Representations}.

\bibitem[{Chi et~al.(2023)Chi, Fan, Rudnicky, and Ramadge}]{chi2023dissecting}
Ta-Chung Chi, Ting-Han Fan, Alexander Rudnicky, and Peter Ramadge. 2023.
\newblock Dissecting transformer length extrapolation via the lens of receptive
  field analysis.
\newblock In \emph{Proceedings of the 61st Annual Meeting of the Association
  for Computational Linguistics (Volume 1: Long Papers)}, pages 13522--13537.

\bibitem[{Dai et~al.(2019)Dai, Yang, Yang, Carbonell, Le, and
  Salakhutdinov}]{dai2019Transformer}
Zihang Dai, Zhilin Yang, Yiming Yang, Jaime~G Carbonell, Quoc Le, and Ruslan
  Salakhutdinov. 2019.
\newblock Transformer-xl: Attentive language models beyond a fixed-length
  context.
\newblock In \emph{Proceedings of the 57th Annual Meeting of the Association
  for Computational Linguistics}, pages 2978--2988.

\bibitem[{Dasigi et~al.(2021)Dasigi, Lo, Beltagy, Cohan, Smith, and
  Gardner}]{dasigi2021dataset}
Pradeep Dasigi, Kyle Lo, Iz~Beltagy, Arman Cohan, Noah~A Smith, and Matt
  Gardner. 2021.
\newblock A dataset of information-seeking questions and answers anchored in
  research papers.
\newblock In \emph{Proceedings of the 2021 Conference of the North American
  Chapter of the Association for Computational Linguistics: Human Language
  Technologies}, pages 4599--4610.

\bibitem[{Ding et~al.(2023)Ding, Ma, Dong, Zhang, Huang, Wang, and
  Wei}]{ding2023longnet}
Jiayu Ding, Shuming Ma, Li~Dong, Xingxing Zhang, Shaohan Huang, Wenhui Wang,
  and Furu Wei. 2023.
\newblock Longnet: Scaling transformers to 1,000,000,000 tokens.
\newblock \emph{arXiv preprint arXiv:2307.02486}.

\bibitem[{Ding et~al.(2024)Ding, Zhang, Zhang, Xu, Shang, Xu, Yang, and
  Yang}]{ding2024longrope}
Yiran Ding, Li~Lyna Zhang, Chengruidong Zhang, Yuanyuan Xu, Ning Shang, Jiahang
  Xu, Fan Yang, and Mao Yang. 2024.
\newblock Longrope: Extending llm context window beyond 2 million tokens.
\newblock \emph{arXiv preprint arXiv:2402.13753}.

\bibitem[{Dong et~al.(2024)Dong, Yang, Zhang, Wang, Chi, and
  Chen}]{dong2024get}
Harry Dong, Xinyu Yang, Zhenyu Zhang, Zhangyang Wang, Yuejie Chi, and Beidi
  Chen. 2024.
\newblock Get more with less: Synthesizing recurrence with kv cache compression
  for efficient llm inference.
\newblock \emph{arXiv preprint arXiv:2402.09398}.

\bibitem[{Gao et~al.(2020)Gao, Biderman, Black, Golding, Hoppe, Foster, Phang,
  He, Thite, Nabeshima, Presser, and Leahy}]{pile}
Leo Gao, Stella Biderman, Sid Black, Laurence Golding, Travis Hoppe, Charles
  Foster, Jason Phang, Horace He, Anish Thite, Noa Nabeshima, Shawn Presser,
  and Connor Leahy. 2020.
\newblock The {P}ile: An 800gb dataset of diverse text for language modeling.
\newblock \emph{arXiv preprint arXiv:2101.00027}.

\bibitem[{Guu et~al.(2020)Guu, Lee, Tung, Pasupat, and
  Chang}]{guu2020retrieval}
Kelvin Guu, Kenton Lee, Zora Tung, Panupong Pasupat, and Mingwei Chang. 2020.
\newblock Retrieval augmented language model pre-training.
\newblock In \emph{International conference on machine learning}, pages
  3929--3938. PMLR.

\bibitem[{Haussler(2018)}]{haussler2018decision}
David Haussler. 2018.
\newblock Decision theoretic generalizations of the pac model for neural net
  and other learning applications.
\newblock In \emph{The Mathematics of Generalization}, pages 37--116. CRC
  Press.

\bibitem[{Huang et~al.(2023)Huang, Zhang, Cheng, and Yang}]{huang2023boosting}
Xijie Huang, Li~Lyna Zhang, Kwang-Ting Cheng, and Mao Yang. 2023.
\newblock Boosting llm reasoning: Push the limits of few-shot learning with
  reinforced in-context pruning.
\newblock \emph{arXiv preprint arXiv:2312.08901}.

\bibitem[{Jiang et~al.(2023)Jiang, Wu, Luo, Li, Lin, Yang, and
  Qiu}]{jiang2023longllmlingua}
Huiqiang Jiang, Qianhui Wu, Xufang Luo, Dongsheng Li, Chin-Yew Lin, Yuqing
  Yang, and Lili Qiu. 2023.
\newblock Longllmlingua: Accelerating and enhancing llms in long context
  scenarios via prompt compression.
\newblock \emph{arXiv preprint arXiv:2310.06839}.

\bibitem[{Kaiokendev(2023)}]{superhot}
Kaiokendev. 2023.
\newblock Things i\'m learning while training superhot.
\newblock \url{https://kaiokendev.github.io/til#extending-context-to-8k}.

\bibitem[{Kaiser et~al.(2016)Kaiser, Nachum, Roy, and
  Bengio}]{kaiser2016learning}
Lukasz Kaiser, Ofir Nachum, Aurko Roy, and Samy Bengio. 2016.
\newblock Learning to remember rare events.
\newblock In \emph{International Conference on Learning Representations}.

\bibitem[{Kazemnejad et~al.(2023)Kazemnejad, Padhi, Ramamurthy, Das, and
  Reddy}]{kazemnejad2023impact}
Amirhossein Kazemnejad, Inkit Padhi, Karthikeyan~Natesan Ramamurthy, Payel Das,
  and Siva Reddy. 2023.
\newblock The impact of positional encoding on length generalization in
  transformers.
\newblock \emph{arXiv preprint arXiv:2305.19466}.

\bibitem[{Ke et~al.(2020)Ke, He, and Liu}]{ke2020rethinking}
Guolin Ke, Di~He, and Tie-Yan Liu. 2020.
\newblock Rethinking positional encoding in language pre-training.
\newblock In \emph{International Conference on Learning Representations}.

\bibitem[{Kenton and Toutanova(2019)}]{kenton2019bert}
Jacob Devlin Ming-Wei~Chang Kenton and Lee~Kristina Toutanova. 2019.
\newblock Bert: Pre-training of deep bidirectional transformers for language
  understanding.
\newblock In \emph{Proceedings of NAACL-HLT}, pages 4171--4186.

\bibitem[{Khandelwal et~al.(2019)Khandelwal, Levy, Jurafsky, Zettlemoyer, and
  Lewis}]{khandelwal2019generalization}
Urvashi Khandelwal, Omer Levy, Dan Jurafsky, Luke Zettlemoyer, and Mike Lewis.
  2019.
\newblock Generalization through memorization: Nearest neighbor language
  models.
\newblock In \emph{International Conference on Learning Representations}.

\bibitem[{Kiyono et~al.(2021)Kiyono, Kobayashi, Suzuki, and
  Inui}]{kiyono2021shape}
Shun Kiyono, Sosuke Kobayashi, Jun Suzuki, and Kentaro Inui. 2021.
\newblock Shape: Shifted absolute position embedding for transformers.
\newblock In \emph{Proceedings of the 2021 Conference on Empirical Methods in
  Natural Language Processing}, pages 3309--3321.

\bibitem[{Lee et~al.(2024)Lee, Chen, Furuta, Canny, and Fischer}]{lee2024human}
Kuang-Huei Lee, Xinyun Chen, Hiroki Furuta, John Canny, and Ian Fischer. 2024.
\newblock A human-inspired reading agent with gist memory of very long
  contexts.
\newblock \emph{arXiv preprint arXiv:2402.09727}.

\bibitem[{Li et~al.(2023)Li, Shao, Xie, Sheng, Zheng, Gonzalez, Stoica, Ma, and
  Zhang}]{longchat2023}
Dacheng Li, Rulin Shao, Anze Xie, Ying Sheng, Lianmin Zheng, Joseph~E.
  Gonzalez, Ion Stoica, Xuezhe Ma, and Hao Zhang. 2023.
\newblock \href {https://lmsys.org/blog/2023-06-29-longchat} {How long can
  open-source llms truly promise on context length?}

\bibitem[{Likhomanenko et~al.(2021)Likhomanenko, Xu, Synnaeve, Collobert, and
  Rogozhnikov}]{likhomanenko2021cape}
Tatiana Likhomanenko, Qiantong Xu, Gabriel Synnaeve, Ronan Collobert, and Alex
  Rogozhnikov. 2021.
\newblock Cape: Encoding relative positions with continuous augmented
  positional embeddings.
\newblock \emph{Advances in Neural Information Processing Systems},
  34:16079--16092.

\bibitem[{Lin(2004)}]{lin2004rouge}
Chin-Yew Lin. 2004.
\newblock Rouge: A package for automatic evaluation of summaries.
\newblock In \emph{Text summarization branches out}, pages 74--81.

\bibitem[{Liu et~al.(2023)Liu, Yan, An, Qiu, and Lin}]{liu2023scaling}
Xiaoran Liu, Hang Yan, Chenxin An, Xipeng Qiu, and Dahua Lin. 2023.
\newblock Scaling laws of rope-based extrapolation.
\newblock In \emph{The Twelfth International Conference on Learning
  Representations}.

\bibitem[{Luo et~al.(2024)Luo, Liu, Xiao, and Liu}]{luo2024bge}
Kun Luo, Zheng Liu, Shitao Xiao, and Kang Liu. 2024.
\newblock Bge landmark embedding: A chunking-free embedding method for
  retrieval augmented long-context large language models.
\newblock \emph{arXiv preprint arXiv:2402.11573}.

\bibitem[{Lv et~al.(2024)Lv, Liu, Guo, Yan, He, Qiu, and
  Lin}]{lv2024longwanjuan}
Kai Lv, Xiaoran Liu, Qipeng Guo, Hang Yan, Conghui He, Xipeng Qiu, and Dahua
  Lin. 2024.
\newblock Longwanjuan: Towards systematic measurement for long text quality.
\newblock \emph{arXiv preprint arXiv:2402.13583}.

\bibitem[{Mohtashami and Jaggi(2023)}]{mohtashami2023landmark}
Amirkeivan Mohtashami and Martin Jaggi. 2023.
\newblock Landmark attention: Random-access infinite context length for
  transformers.
\newblock \emph{arXiv preprint arXiv:2305.16300}.

\bibitem[{Oren et~al.(2024)Oren, Hassid, Adi, and
  Schwartz}]{oren2024transformers}
Matanel Oren, Michael Hassid, Yossi Adi, and Roy Schwartz. 2024.
\newblock Transformers are multi-state rnns.
\newblock \emph{arXiv preprint arXiv:2401.06104}.

\bibitem[{Papineni et~al.(2002)Papineni, Roukos, Ward, and
  Zhu}]{papineni2002bleu}
Kishore Papineni, Salim Roukos, Todd Ward, and Wei-Jing Zhu. 2002.
\newblock Bleu: a method for automatic evaluation of machine translation.
\newblock In \emph{Proceedings of the 40th annual meeting of the Association
  for Computational Linguistics}, pages 311--318.

\bibitem[{Peng et~al.(2023)Peng, Quesnelle, Fan, and Shippole}]{peng2023yarn}
Bowen Peng, Jeffrey Quesnelle, Honglu Fan, and Enrico Shippole. 2023.
\newblock Yarn: Efficient context window extension of large language models.
\newblock In \emph{The Twelfth International Conference on Learning
  Representations}.

\bibitem[{Peysakhovich and Lerer(2023)}]{peysakhovich2023attention}
Alexander Peysakhovich and Adam Lerer. 2023.
\newblock Attention sorting combats recency bias in long context language
  models.
\newblock \emph{arXiv preprint arXiv:2310.01427}.

\bibitem[{Pollard(1990)}]{pollard1990empirical}
David Pollard. 1990.
\newblock Empirical processes: Theory and applications.
\newblock In \emph{NSF-CBMS Regional Conference Series in Probability and
  Statistics}, pages i--86. JSTOR.

\bibitem[{Press et~al.(2021)Press, Smith, and Lewis}]{press2021train}
Ofir Press, Noah Smith, and Mike Lewis. 2021.
\newblock Train short, test long: Attention with linear biases enables input
  length extrapolation.
\newblock In \emph{International Conference on Learning Representations}.

\bibitem[{Qiu et~al.(2024)Qiu, Li, Huang, Zhong, and King}]{qiu2024clongeval}
Zexuan Qiu, Jingjing Li, Shijue Huang, Wanjun Zhong, and Irwin King. 2024.
\newblock Clongeval: A chinese benchmark for evaluating long-context large
  language models.
\newblock \emph{arXiv preprint arXiv:2403.03514}.

\bibitem[{Qu(2023)}]{qu2023gpt}
Zhijie Qu. 2023.
\newblock Gpt rotational position embedding for length extrapolation.
\newblock In \emph{Proceedings of the 2023 6th International Conference on
  Machine Learning and Natural Language Processing}, pages 166--170.

\bibitem[{Raffel et~al.(2020)Raffel, Shazeer, Roberts, Lee, Narang, Matena,
  Zhou, Li, and Liu}]{raffel2020exploring}
Colin Raffel, Noam Shazeer, Adam Roberts, Katherine Lee, Sharan Narang, Michael
  Matena, Yanqi Zhou, Wei Li, and Peter~J Liu. 2020.
\newblock Exploring the limits of transfer learning with a unified text-to-text
  transformer.
\newblock \emph{The Journal of Machine Learning Research}, 21(1):5485--5551.

\bibitem[{Rasley et~al.(2020)Rasley, Rajbhandari, Ruwase, and
  He}]{rasley2020deepspeed}
Jeff Rasley, Samyam Rajbhandari, Olatunji Ruwase, and Yuxiong He. 2020.
\newblock Deepspeed: System optimizations enable training deep learning models
  with over 100 billion parameters.
\newblock In \emph{Proceedings of the 26th ACM SIGKDD International Conference
  on Knowledge Discovery \& Data Mining}, pages 3505--3506.

\bibitem[{Ratner et~al.(2023)Ratner, Levine, Belinkov, Ram, Magar, Abend,
  Karpas, Shashua, Leyton-Brown, and Shoham}]{ratner2023parallel}
Nir Ratner, Yoav Levine, Yonatan Belinkov, Ori Ram, Inbal Magar, Omri Abend,
  Ehud Karpas, Amnon Shashua, Kevin Leyton-Brown, and Yoav Shoham. 2023.
\newblock Parallel context windows for large language models.
\newblock In \emph{Proceedings of the 61st Annual Meeting of the Association
  for Computational Linguistics (Volume 1: Long Papers)}, pages 6383--6402.

\bibitem[{Ren and Zhu(2024)}]{ren2024efficacy}
Siyu Ren and Kenny~Q Zhu. 2024.
\newblock On the efficacy of eviction policy for key-value constrained
  generative language model inference.
\newblock \emph{arXiv preprint arXiv:2402.06262}.

\bibitem[{Shao et~al.(2024)Shao, Xiao, Liu, and Zhang}]{shao2024flexibly}
Ninglu Shao, Shitao Xiao, Zheng Liu, and Peitian Zhang. 2024.
\newblock Flexibly scaling large language models contexts through extensible
  tokenization.
\newblock \emph{arXiv preprint arXiv:2401.07793}.

\bibitem[{Song et~al.(2023)Song, Wang, Cho, Pan, and Yu}]{song2023zebra}
Kaiqiang Song, Xiaoyang Wang, Sangwoo Cho, Xiaoman Pan, and Dong Yu. 2023.
\newblock Zebra: Extending context window with layerwise grouped local-global
  attention.
\newblock \emph{arXiv preprint arXiv:2312.08618}.

\bibitem[{Su et~al.(2021)Su, Lu, Pan, Murtadha, Wen, and Liu}]{su2021roformer}
Jianlin Su, Yu~Lu, Shengfeng Pan, Ahmed Murtadha, Bo~Wen, and Yunfeng Liu.
  2021.
\newblock Roformer: Enhanced transformer with rotary position embedding.
\newblock \emph{arXiv preprint arXiv:2104.09864}.

\bibitem[{Sun et~al.(2022)Sun, Dong, Patra, Ma, Huang, Benhaim, Chaudhary,
  Song, and Wei}]{sun2022length}
Yutao Sun, Li~Dong, Barun Patra, Shuming Ma, Shaohan Huang, Alon Benhaim,
  Vishrav Chaudhary, Xia Song, and Furu Wei. 2022.
\newblock A length-extrapolatable transformer.
\newblock \emph{arXiv preprint arXiv:2212.10554}.

\bibitem[{Tao et~al.(2023)Tao, Feng, and Zhao}]{tao2023frustratingly}
Mingxu Tao, Yansong Feng, and Dongyan Zhao. 2023.
\newblock A frustratingly easy improvement for position embeddings via random
  padding.
\newblock \emph{arXiv preprint arXiv:2305.04859}.

\bibitem[{Team(2023)}]{MosaicML2023Introducing}
MosaicML~NLP Team. 2023.
\newblock \href {www.mosaicml.com/blog/mpt-7b} {Introducing mpt-7b: A new
  standard for open-source, commercially usable llms}.
\newblock Accessed: 2023-05-05.

\bibitem[{Touvron et~al.(2023{\natexlab{a}})Touvron, Lavril, Izacard, Martinet,
  Lachaux, Lacroix, Rozi{\`e}re, Goyal, Hambro, Azhar
  et~al.}]{touvron2023llama}
Hugo Touvron, Thibaut Lavril, Gautier Izacard, Xavier Martinet, Marie-Anne
  Lachaux, Timoth{\'e}e Lacroix, Baptiste Rozi{\`e}re, Naman Goyal, Eric
  Hambro, Faisal Azhar, et~al. 2023{\natexlab{a}}.
\newblock Llama: Open and efficient foundation language models.
\newblock \emph{arXiv preprint arXiv:2302.13971}.

\bibitem[{Touvron et~al.(2023{\natexlab{b}})Touvron, Martin, Stone, Albert,
  Almahairi, Babaei, Bashlykov, Batra, Bhargava, Bhosale
  et~al.}]{touvron2023llama2}
Hugo Touvron, Louis Martin, Kevin Stone, Peter Albert, Amjad Almahairi, Yasmine
  Babaei, Nikolay Bashlykov, Soumya Batra, Prajjwal Bhargava, Shruti Bhosale,
  et~al. 2023{\natexlab{b}}.
\newblock Llama 2: Open foundation and fine-tuned chat models.
\newblock \emph{arXiv preprint arXiv:2307.09288}.

\bibitem[{Tworkowski et~al.(2023)Tworkowski, Staniszewski, Pacek, Wu,
  Michalewski, and Mi{\l}o{\'s}}]{tworkowski2023focused}
Szymon Tworkowski, Konrad Staniszewski, Miko{\l}aj Pacek, Yuhuai Wu, Henryk
  Michalewski, and Piotr Mi{\l}o{\'s}. 2023.
\newblock Focused transformer: Contrastive training for context scaling.
\newblock \emph{arXiv preprint arXiv:2307.03170}.

\bibitem[{Vaswani et~al.(2017)Vaswani, Shazeer, Parmar, Uszkoreit, Jones,
  Gomez, Kaiser, and Polosukhin}]{vaswani2017attention}
Ashish Vaswani, Noam Shazeer, Niki Parmar, Jakob Uszkoreit, Llion Jones,
  Aidan~N Gomez, {\L}ukasz Kaiser, and Illia Polosukhin. 2017.
\newblock Attention is all you need.
\newblock \emph{Advances in neural information processing systems}, 30.

\bibitem[{Wang and Komatsuzaki(2021)}]{gpt-j}
Ben Wang and Aran Komatsuzaki. 2021.
\newblock {GPT-J-6B: A 6 Billion Parameter Autoregressive Language Model}.
\newblock \url{https://github.com/kingoflolz/mesh-transformer-jax}.

\bibitem[{Wang et~al.(2024)Wang, Ning, Pan, Wu, Guo, Deng, Bao, Wang, and
  Zhang}]{wang2024novelqa}
Cunxiang Wang, Ruoxi Ning, Boqi Pan, Tonghui Wu, Qipeng Guo, Cheng Deng,
  Guangsheng Bao, Qian Wang, and Yue Zhang. 2024.
\newblock Novelqa: A benchmark for long-range novel question answering.
\newblock \emph{arXiv preprint arXiv:2403.12766}.

\bibitem[{Wu et~al.(2021)Wu, Rabe, Hutchins, and Szegedy}]{wu2021memorizing}
Yuhuai Wu, Markus~Norman Rabe, DeLesley Hutchins, and Christian Szegedy. 2021.
\newblock Memorizing transformers.
\newblock In \emph{International Conference on Learning Representations}.

\bibitem[{Xiao et~al.(2024)Xiao, Tian, Chen, Han, and
  Lewis}]{xiao2024efficient}
Guangxuan Xiao, Yuandong Tian, Beidi Chen, Song Han, and Mike Lewis. 2024.
\newblock \href {https://openreview.net/forum?id=NG7sS51zVF} {Efficient
  streaming language models with attention sinks}.
\newblock In \emph{The Twelfth International Conference on Learning
  Representations}.

\bibitem[{Yang et~al.(2023)Yang, Klein, Peng, and Tian}]{yang-etal-2023-doc}
Kevin Yang, Dan Klein, Nanyun Peng, and Yuandong Tian. 2023.
\newblock \href {https://doi.org/10.18653/v1/2023.acl-long.190} {{DOC}:
  Improving long story coherence with detailed outline control}.
\newblock In \emph{Proceedings of the 61st Annual Meeting of the Association
  for Computational Linguistics (Volume 1: Long Papers)}, pages 3378--3465,
  Toronto, Canada. Association for Computational Linguistics.

\bibitem[{Yang and Hua(2024)}]{yang2024attendre}
Zi~Yang and Nan Hua. 2024.
\newblock Attendre: Wait to attend by retrieval with evicted queries in
  memory-based transformers for long context processing.
\newblock \emph{arXiv preprint arXiv:2401.04881}.

\bibitem[{Yogatama et~al.(2021)Yogatama, de~Masson~d’Autume, and
  Kong}]{yogatama2021adaptive}
Dani Yogatama, Cyprien de~Masson~d’Autume, and Lingpeng Kong. 2021.
\newblock Adaptive semiparametric language models.
\newblock \emph{Transactions of the Association for Computational Linguistics},
  9:362--373.

\bibitem[{Yuan et~al.(2024)Yuan, Ning, Zhou, Yang, Li, Zhuang, Tan, Yao, Lin,
  Li et~al.}]{yuan2024lv}
Tao Yuan, Xuefei Ning, Dong Zhou, Zhijie Yang, Shiyao Li, Minghui Zhuang,
  Zheyue Tan, Zhuyu Yao, Dahua Lin, Boxun Li, et~al. 2024.
\newblock Lv-eval: A balanced long-context benchmark with 5 length levels up to
  256k.
\newblock \emph{arXiv preprint arXiv:2402.05136}.

\bibitem[{Zaheer et~al.(2020)Zaheer, Guruganesh, Dubey, Ainslie, Alberti,
  Ontanon, Pham, Ravula, Wang, Yang et~al.}]{zaheer2020big}
Manzil Zaheer, Guru Guruganesh, Kumar~Avinava Dubey, Joshua Ainslie, Chris
  Alberti, Santiago Ontanon, Philip Pham, Anirudh Ravula, Qifan Wang, Li~Yang,
  et~al. 2020.
\newblock Big bird: Transformers for longer sequences.
\newblock \emph{Advances in neural information processing systems},
  33:17283--17297.

\bibitem[{Zhang et~al.(2024{\natexlab{a}})Zhang, Liu, Xiao, Shao, Ye, and
  Dou}]{zhang2024soaring}
Peitian Zhang, Zheng Liu, Shitao Xiao, Ninglu Shao, Qiwei Ye, and Zhicheng Dou.
  2024{\natexlab{a}}.
\newblock Soaring from 4k to 400k: Extending llm's context with activation
  beacon.
\newblock \emph{arXiv preprint arXiv:2401.03462}.

\bibitem[{Zhang et~al.(2024{\natexlab{b}})Zhang, Chen, Hu, Xu, Chen, Hao, Han,
  Thai, Wang, Liu et~al.}]{zhang2024infty}
Xinrong Zhang, Yingfa Chen, Shengding Hu, Zihang Xu, Junhao Chen, Moo~Khai Hao,
  Xu~Han, Zhen~Leng Thai, Shuo Wang, Zhiyuan Liu, et~al. 2024{\natexlab{b}}.
\newblock $\infty$ bench: Extending long context evaluation beyond 100k tokens.
\newblock \emph{arXiv preprint arXiv:2402.13718}.

\bibitem[{Zhang et~al.(2024{\natexlab{c}})Zhang, Chen, Liu, Yao, Ruwase, Chen,
  Wu, and Wang}]{zhang2024found}
Zhenyu Zhang, Runjin Chen, Shiwei Liu, Zhewei Yao, Olatunji Ruwase, Beidi Chen,
  Xiaoxia Wu, and Zhangyang Wang. 2024{\natexlab{c}}.
\newblock Found in the middle: How language models use long contexts better via
  plug-and-play positional encoding.
\newblock \emph{arXiv preprint arXiv:2403.04797}.

\bibitem[{Zhang et~al.(2024{\natexlab{d}})Zhang, Sheng, Zhou, Chen, Zheng, Cai,
  Song, Tian, R{\'e}, Barrett et~al.}]{zhang2024h2o}
Zhenyu Zhang, Ying Sheng, Tianyi Zhou, Tianlong Chen, Lianmin Zheng, Ruisi Cai,
  Zhao Song, Yuandong Tian, Christopher R{\'e}, Clark Barrett, et~al.
  2024{\natexlab{d}}.
\newblock H2o: Heavy-hitter oracle for efficient generative inference of large
  language models.
\newblock \emph{Advances in Neural Information Processing Systems}, 36.

\bibitem[{Zhou et~al.(2023)Zhou, Jiang, Cui, Wang, Xiao, Hou, Cotterell, and
  Sachan}]{zhou2023recurrentgpt}
Wangchunshu Zhou, Yuchen~Eleanor Jiang, Peng Cui, Tiannan Wang, Zhenxin Xiao,
  Yifan Hou, Ryan Cotterell, and Mrinmaya Sachan. 2023.
\newblock Recurrentgpt: Interactive generation of (arbitrarily) long text.
\newblock \emph{arXiv preprint arXiv:2305.13304}.

\bibitem[{Zhu et~al.(2023)Zhu, Yang, Wang, Song, Wu, Wei, and Li}]{zhu2023pose}
Dawei Zhu, Nan Yang, Liang Wang, Yifan Song, Wenhao Wu, Furu Wei, and Sujian
  Li. 2023.
\newblock Pose: Efficient context window extension of llms via positional
  skip-wise training.
\newblock In \emph{The Twelfth International Conference on Learning
  Representations}.

\end{thebibliography}
